\documentclass[10pt,conference]{IEEEtran}
\IEEEoverridecommandlockouts
\usepackage{cite}
\usepackage{amsmath,amssymb,amsfonts}
\usepackage{textcomp}
\usepackage[table]{xcolor}
\usepackage{colortbl}
\usepackage{algorithm}
\usepackage{algpseudocode}
\usepackage{graphicx}

\usepackage{booktabs}
\newcommand \ignore[1]{}
\usepackage{textcomp}
\usepackage{xcolor}
\usepackage{multirow}
\usepackage{makecell}
\usepackage{subcaption}

\usepackage[most]{tcolorbox}

\usepackage{colortbl}

\definecolor{paretoyellow}{RGB}{252,237,140}

\usepackage{caption}
\def\BibTeX{{\rm B\kern-.05em{\sc i\kern-.025em b}\kern-.08em
    T\kern-.1667em\lower.7ex\hbox{E}\kern-.125emX}}
\begin{document}

\title{Rethinking Generative Reconstruction Attacks against Graph Neural Network Models

\thanks{$\dagger$ Corresponding author.}
}

\author{\IEEEauthorblockN{Adebayo Keji\IEEEauthorrefmark{4},
Sayanton Dibbo $\dagger$\IEEEauthorrefmark{5}
}
 
\IEEEauthorblockA{Department of Computer Science,
The University of Alabama\\
aakeji@crimson.ua.edu\IEEEauthorrefmark{4},
sdibbo@ua.edu\IEEEauthorrefmark{5}}
}

\ignore{\author{\IEEEauthorblockN{1\textsuperscript{st} Given Name Surname}
\IEEEauthorblockA{\textit{dept. name of organization (of Aff.)} \\
\textit{name of organization (of Aff.)}\\
City, Country 
email address or ORCID}
\and
\IEEEauthorblockN{2\textsuperscript{nd} Given Name Surname}
\IEEEauthorblockA{\textit{dept. name of organization (of Aff.)} \\
\textit{name of organization (of Aff.)}\\
City, Country \\
email address or ORCID}
\and
\IEEEauthorblockN{3\textsuperscript{rd} Given Name Surname}
\IEEEauthorblockA{\textit{dept. name of organization (of Aff.)} \\
\textit{name of organization (of Aff.)}\\
City, Country \\
email address or ORCID}
\and
\IEEEauthorblockN{4\textsuperscript{th} Given Name Surname}
\IEEEauthorblockA{\textit{dept. name of organization (of Aff.)} \\
\textit{name of organization (of Aff.)}\\
City, Country \\
email address or ORCID}
\and
\IEEEauthorblockN{5\textsuperscript{th} Given Name Surname}
\IEEEauthorblockA{\textit{dept. name of organization (of Aff.)} \\
\textit{name of organization (of Aff.)}\\
City, Country \\
email address or ORCID}
\and
\IEEEauthorblockN{6\textsuperscript{th} Given Name Surname}
\IEEEauthorblockA{\textit{dept. name of organization (of Aff.)} \\
\textit{name of organization (of Aff.)}\\
City, Country \\
email address or ORCID}
}}
\maketitle

\begin{abstract}
\textcolor{black}{The application of graph data in numerous disciplines raises the need for gathering and analyzing huge volumes of data, some of which is private and sensitive. The non-Euclidean nature of the graph data makes the analysis computationally challenging, leading to the use of Graph Neural Networks (GNNs) in the age of AI. GNNs may inadvertently leak sensitive data they are trained on, which raises serious data security issues, including the model inversion attack. In this study, we analyze GNNs' vulnerabilities by introducing two novel graph inversion (i.e., reconstruction) attacks: \textcolor{black}{graph-label conditioned (GLC)} attack and \textcolor{black}{embedding-label conditioned (ELC)} attack, utilizing target-model predictions and their intermediate representations, respectively. We perform a comprehensive analysis of our introduced privacy attacks and compare them with existing baselines across three benchmark graph datasets (i.e., NCI1, PROTEINS, and AIDS) and four graph distributional/structural metrics (i.e., FGD, EGD, MMD, and GKS). Our work demonstrates that an adversary can use the generator-discriminator technique to reconstruct high-quality graphs in real-world \textit{black-box} attack scenarios against GNNs. Additionally, we present a variant of our attacks (Ours$--$) with 50\% reduced queries, achieving good or comparable reconstruction attack performance. In addition, we show that GNNs are highly vulnerable to privacy attacks, varying \textit{Laplacian} noise-scales.} 
\end{abstract}


\section{Introduction}
\label{sec:intro}

Graph data have recently gained momentum in their application across numerous domains, which include chemistry and drug discovery~\cite{zeng2022toward,gaudelet2021utilizing}, social network analysis~\cite{fan2012graph}, supply chain and logistics~\cite{kosasih2022machine}, and financial networks~\cite{hoang2023machine}. The widespread use of artificial intelligence (AI) assisted by machine learning (ML) models to address a wide range of real-world issues necessitated the collection and processing of massive amounts of data, some of which were sensitive and personal, posing significant data protection concerns~\cite{el2024preserving, dibbo2026understanding}. The main issue with data privacy in ML applications is that models may unintentionally disclose information about the specific data points they were trained on. The application of graph neural networks (GNN) for structured graph analysis and modeling is an emerging field~\cite{he2021overview,janssen2022adoption} that is currently adopted by organizations~\cite{jain2023opinion}. \textcolor{black}{The increased popularity of the GNN has led to a couple of exploitations from attackers to infer sensitive information from the GNN model through inversion~\cite{zhang2021graphmi,zhou2024model} or cause the model to misclassify via adversarial attack~\cite{zugner2020adversarial,dai2018adversarial}}. 

Over the years, credible model inversion (MI) attacks have been carried out, one of which is on linear models~\cite{fredrikson2014privacy,hidano2017model,dibbo2023sok}. Fredrikson et al.~\cite{fredrikson2014privacy} study represents MI's first application, where sensitive patient data were inferred when given the model and patient demographic information. Recent works of MI on DNN~\cite{zhang2020secret,chen2020improved,nguyen2023re,guo2024newfederatedlearningframework} using image data have gathered remarkable contributions. Extension of MI attack on graph data using GNN models~\cite{zhang2022model,he2019model,zhang2021inferenceattacksgraphneural} is receiving considerable attention. The primary focus of this study is to understand the data leakage issue in graph neural networks by conducting a reconstruction attack based on the underlying relationship between model output and its training data.\\
\textbf{Our Contributions:} \textcolor{black}{This research aims to analyze GNNs' underexplored vulnerabilities, i.e., reconstruct sensitive distributional graph data used to train a victim model with access to only model output via model querying, commonly referred to as \textit{black-box} settings. The following contributions were made towards the exploration of data privacy in GNNs.} 
\begin{itemize}
\item \textcolor{black}{We introduce two privacy attacks (GLC and ELC) against GNNs with GAN-based generative methods, one of which involves reconstruction of sensitive data from a victim model using a generative model trained on graph samples and victim model prediction, and the other trained on intermediate representation and true label.}
\item We compare the baseline models (e.g., VAE) privacy attack performances with our introduced attacks on GNNs, varying three commonly used graph datasets (NCI1, PROTEINS, and AIDS) and across four performance metrics (FGD, EGD, MMD, and GKS). Our results indicate that our GAN-based methods in proposed attacks pose higher privacy vulnerabilities to GNNs, even when the adversary has 50\% less query capabilities (i.e., Ours$--$ method).
\item We demonstrate that the proposed reconstruction attacks utilizing generative models (e.g., GAN) outperform other baselines in both attack setups, varying different \textit{Laplacian} noise-scales.
\end{itemize}
\section{Related Work}
\label{sec:preli}
\textcolor{black}{In this section, prior studies in the field of model inversion attacks are highlighted and discussed. Numerous studies~\cite{fredrikson2015model,he2019model,7536387} have been conducted on this attack, including those on CNN~\cite{li2024model,wang2021variational}, GNN~\cite{DBLP:journals/corr/abs-2010-00906,zhang2021graphmi,zhou2024model}, and LLM~\cite{wang2025privacy,qu2025prompt}.} Fredrikson et al.~\cite{10.1145/2810103.2813677} model inversion attack was carried out on a neural network with the exploitation of confidence score value and model prediction to reconstruct its training data. Yang et al~\cite{yang2019adversarialneuralnetworkinversion} train an inversion model without access to the original training data; this work uses the adversary's prior knowledge to create an auxiliary set, then creates a truncation-based method to align the inversion model. An et al.\cite{An2022MIRRORMI} train the inversion model using auxiliary data of the same distribution as the original training data. The authors use a truncation-based method to align the inversion model so that the target model can be effectively inverted from partial prediction. These studies are similar in terms of data used and settings.

The privacy leakage of graph embeddings from GNN models was explored in Duddu et al.~\cite{DBLP:journals/corr/abs-2010-00906}. To reconstruct the target graph, the authors train a graph encoder-decoder attack model using graph data from the same distribution as the target graph data. Zhang et al.~\cite{zhang2021inferenceattacksgraphneural} proposed developing an attack model using a graph encoder-decoder, where both are trained with in-distribution graph data and the decoder is used to reconstruct the target graph data from the available target data embeddings. In Zhan et al.~\cite{10302529}, a graph-decoder attack model is used to reconstruct target graph data from its publicly available embeddings. This work is similar to the study by~\cite{DBLP:journals/corr/abs-2010-00906} with a difference in the computation of GNN. These studies assume access to target graph data embeddings, which can be unrealistic in the real world. Zhang et al.~\cite{zhang2022model} presented a model inversion attack using reinforcement learning and a gradient estimation technique that relies on extremely high queries and estimates the derivative based on error rate, introducing more noise and high time overhead to graph reconstruction.

\section{Generative Adversarial Network }
\label{sec:GAN}

\begin{figure*}[t]
    \centering
    
    \begin{subfigure}[t]{0.88\columnwidth}
        \centering
        \includegraphics[width=\linewidth]{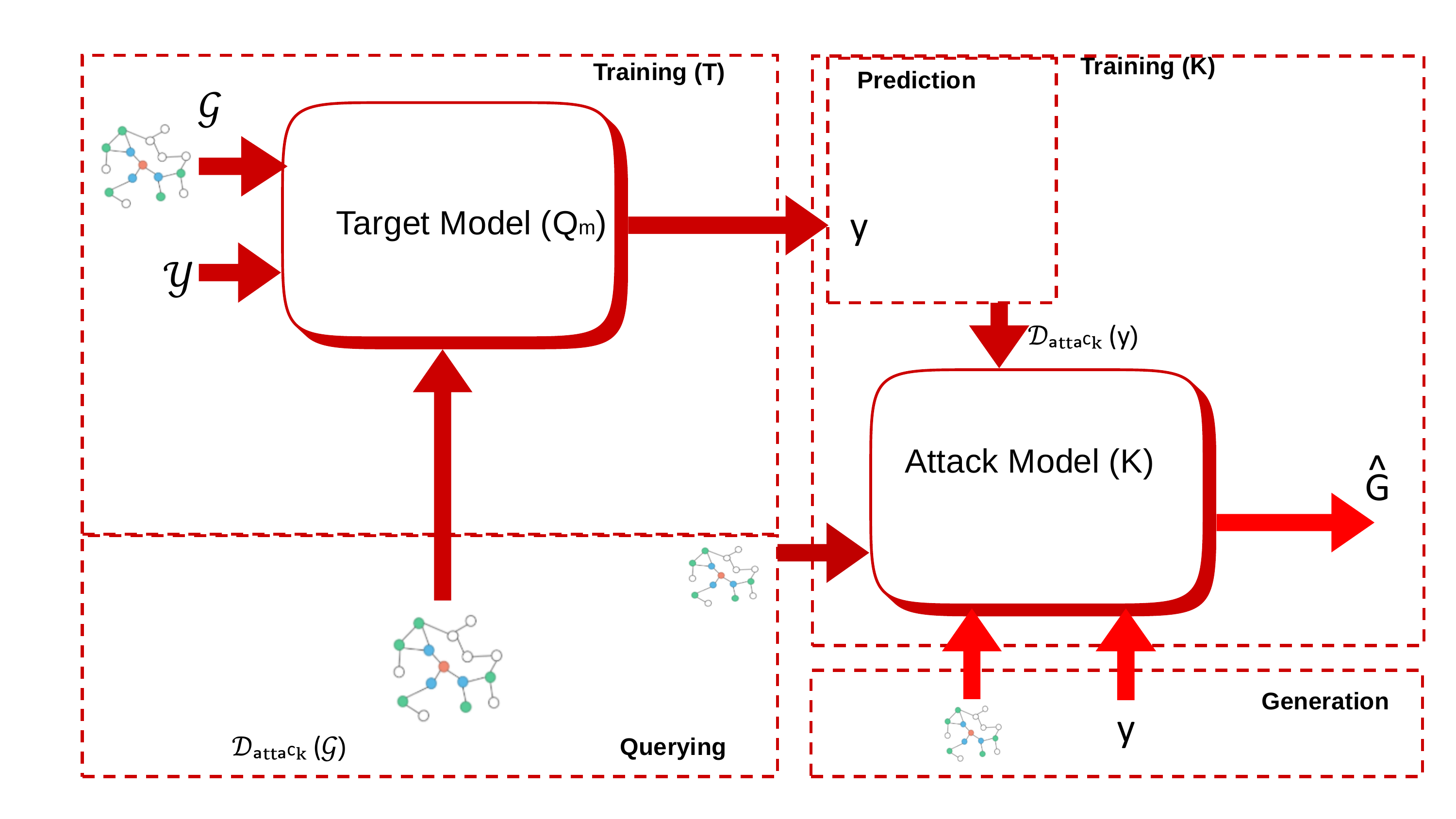}
        \caption{Graph \& Label-Conditioned (GLC) Attack}
        \label{fig:sub1}
    \end{subfigure}
    \begin{subfigure}[t]{0.88\columnwidth}
        \centering
        \includegraphics[width=\linewidth]{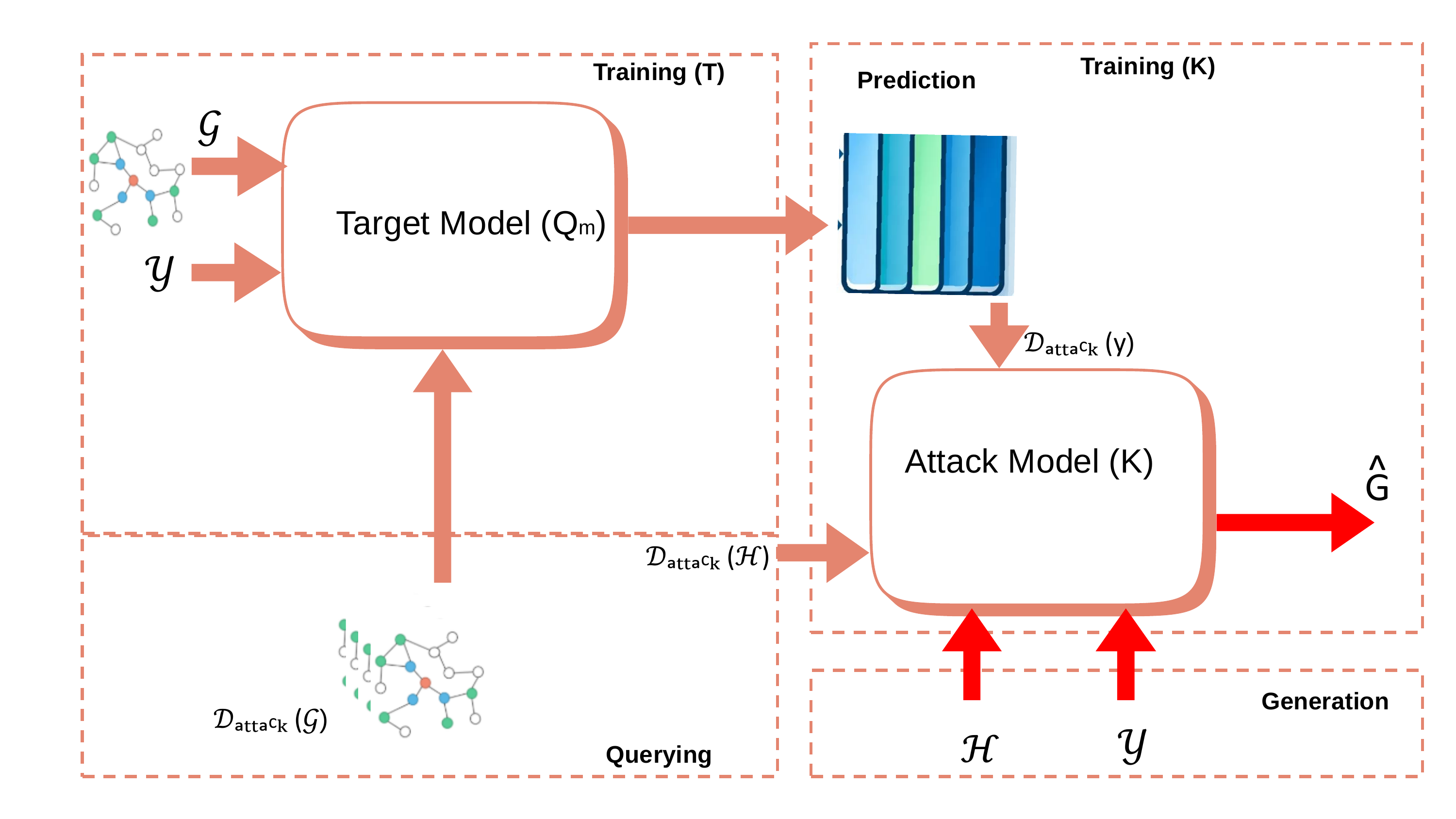}
        \caption{Embedding \& Label-Conditioned (ELC) Attack}
        \label{fig:sub2}
    \end{subfigure}
    
    \caption{An overview of the proposed two novel privacy attacks against graphs: (a) GLC attack and (b) ELC attack.}
    \label{fig:framework}
\end{figure*}

In this section, we define the GAN generative model and explain the application of graph data in GAN-based models.
\subsection{GAN models} 
\textcolor{black}{The Generative Adversarial Network (GAN) model consists of two models, a generative model $S$ and a discriminative model $K$. These networks are trained simultaneously, while the former captures the distribution of the data, the latter estimates the probability of samples originating from the generative model or the data distribution~\cite{10.1145/3422622}.}
The result in equation~\eqref{eqn1} is obtained when $S$ map sample from the true data distribution $x \sim p_{\text{data}}(x)$ and samples latent variable from the prior distribution $z \sim p_z(z)$. 
\begin{equation}
  S: z \rightarrow x' = S(z)
  \label{eqn1}
\end{equation}
where $x'$ is the fake data generated. The probability that a given sample is real is computed by the $K(x)$. However, the GAN objective is defined as a min-max optimization problem. 

\begin{align}
\min_{S} \max_{K} \; V(K, S) =\notag
&\ \mathbb{E}_{x \sim p_{\text{data}}(x)} [\log K(x)] \\ 
&+ \mathbb{E}_{z \sim p_z(z)} [\log (1 - K(S(z)))]
\label{eqn2}
\end{align}

\label{sec:method}

\label{sec:method}

\ignore{
\begin{figure*}[t]
    \centering
    
    \begin{subfigure}[t]{0.88\columnwidth}
        \centering
        \includegraphics[width=\linewidth]{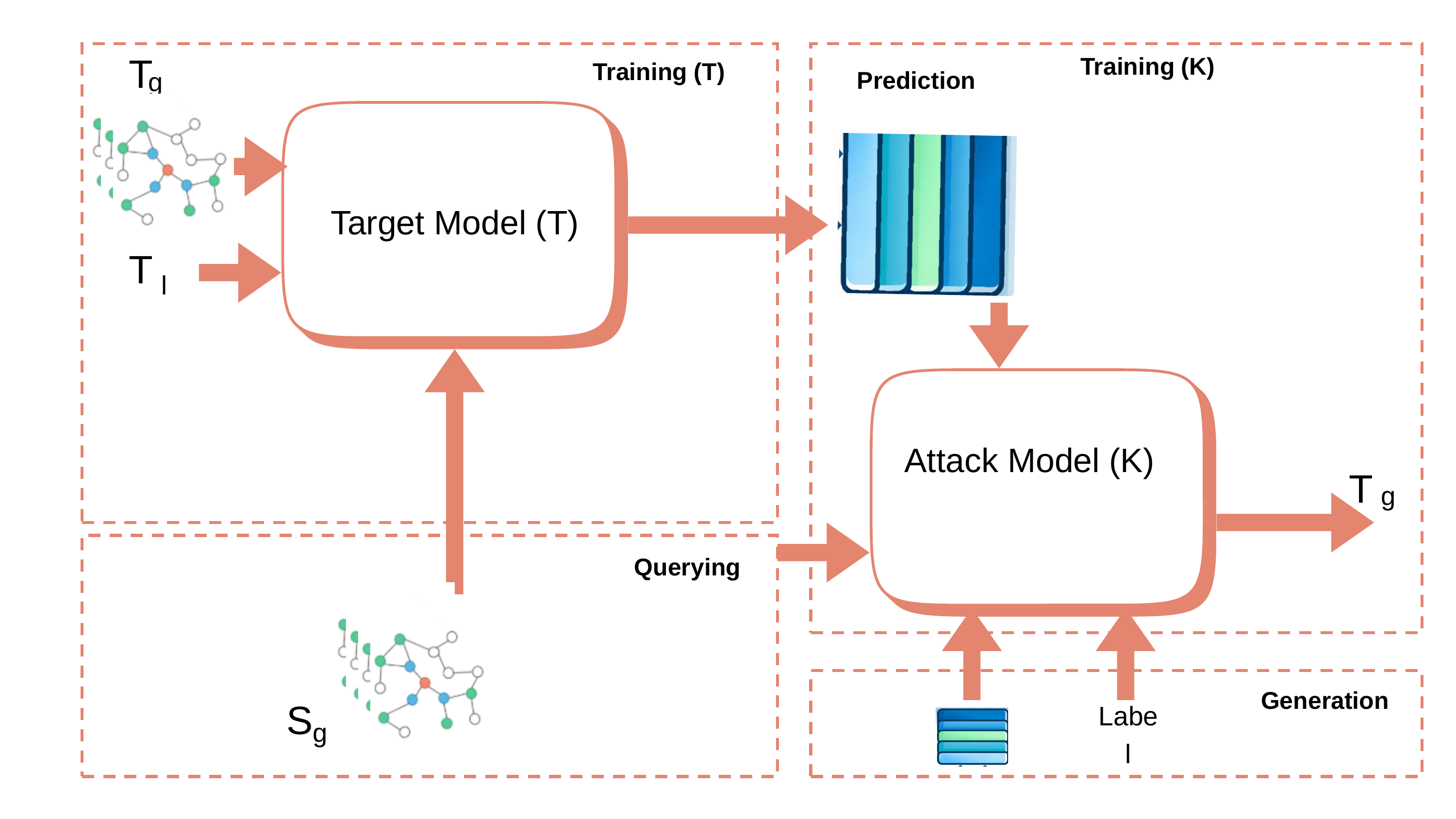}
        \label{fig:sub1}
        \caption{a}
    \end{subfigure}
    \begin{subfigure}[t]{0.88\columnwidth}
        \centering
        \includegraphics[width=\linewidth]{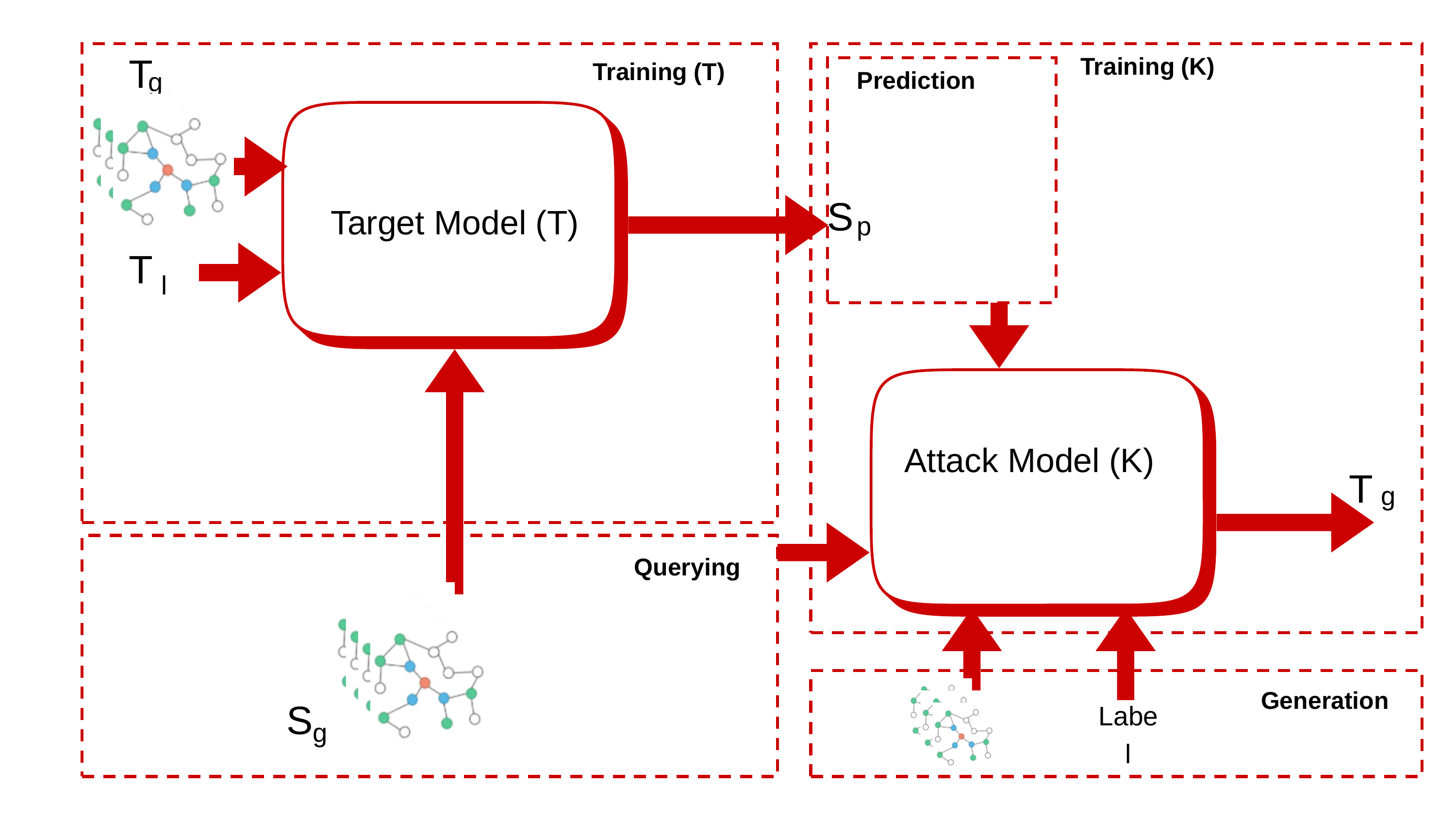}
        \label{fig:sub2}
        \caption{b}
    \end{subfigure}
    
    \caption{A detailed taxonomy of graph-label and embedding-label conditioned reconstruction of graph.}
    \label{fig:framework}
\end{figure*}
}

\subsection{Graph based GAN}
The adoption of GAN for complex graph data structure modeling in this study stems from the ability of the model to apply adversarial training to learn the underlying node relationship and topology of graph data. Let 
$E$ represent the edge set, $\mathcal{V}$ represent the graph node, $\mathcal{G} = (\mathcal{V}, $ E$)$ with node features $\mathbf{X} \in \mathbb{R}^{N \times d}$ and adjacency matrix $\mathbf{A} \in {0,1}^{N \times N}$. The goal is to develop a generative model that represents the distribution $p(\mathbf{A} \mid \mathbf{X})$.\\
\textbf{Model Architecture:}
The GAN architecture consists of a generator $S_{\phi}$ and a discriminator $K_{\psi}$. Given an optional conditioning information like node features and latent noise $\mathbf{z} \sim p_z(\mathbf{z})$, the generation of a synthetic adjacency matrix $\hat{\mathbf{A}}$ is carried out by the generator $S_{\phi}$. 
\begin{equation}
  \hat{\mathbf{A}} = S_{\phi}(\mathbf{z}, \mathbf{X})  
  \label{eqn3}
\end{equation}
Neural architectures that capture paired node interactions are commonly used to create the generator, to ensure that $\hat{\mathbf{A}} \in [0,1]^{N \times N}$. 
Discriminators distinguish between generated graphs and real graphs. It takes as input either $(\hat{\mathbf{A}}, \mathbf{X})$ or $(\mathbf{A}, \mathbf{X})$  and outputs a probability score
$D_{\psi}(\mathbf{A}, \mathbf{X}) \in [0,1]$.\\
\textbf{Adversarial Training Objective:}
The training objective follows a minimax game between $S_{\phi}$ and $K_{\psi}$:

\begin{equation}
\begin{aligned}
\min_{\phi}\max_{\psi}\;
\mathcal{L}_{\mathrm{adv}}
=&\;
\mathbb{E}_{(\mathbf{A},\mathbf{X})\sim p_{\mathrm{data}}}
\left[
\log K_{\psi}(\mathbf{A},\mathbf{X})
\right] \\
&+
\mathbb{E}_{\substack{\mathbf{z}\sim p_z\\
\mathbf{X}\sim p_{\mathrm{data}}}}
\left[
\log\!\left(
1-K_{\psi}
\left(
S_{\phi}(\mathbf{z},\mathbf{X}),
\mathbf{X}
\right)
\right)
\right]
\end{aligned}
\end{equation}
The non-saturation loss is used to optimize the generator to improve the training stability:
\begin{equation}
\mathcal{L}{S} =
\mathbb{E}{\mathbf{z} \sim p_z,\ \mathbf{X} \sim p_{\text{data}}}
\left[ -\log K_{\psi}(S_{\phi}(\mathbf{z}, \mathbf{X}), \mathbf{X}) \right]
\end{equation}
\section{Methods}
\label{sec:method}
\textcolor{black}{In this section, we discuss adversarial goals, assumptions, and capabilities. We also highlight our two proposed attacks and define our threat model.}

\ignore{=======================
\begin{figure*}[!t]
    \centering
    \includegraphics[width=0.7
    \textwidth]{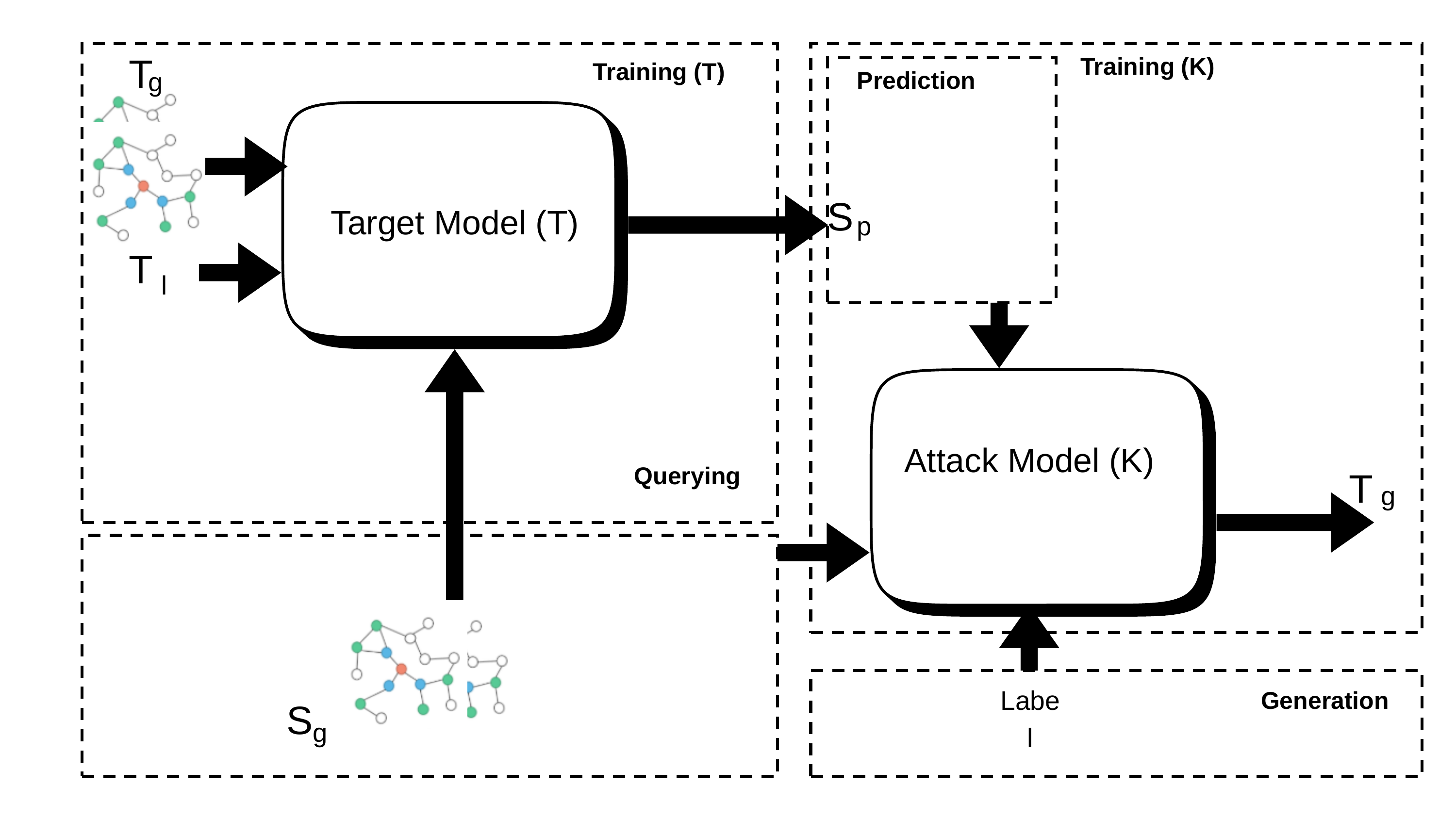}
    \caption{A detailed taxonomy of LOGRA.}
    \label{fig:LOGRA_overview}
\end{figure*}
==============================}

\ignore{=========
\begin{figure*}[t]
    \centering
    
    \begin{subfigure}[t]{0.88\columnwidth}
        \centering
        \includegraphics[width=\linewidth]{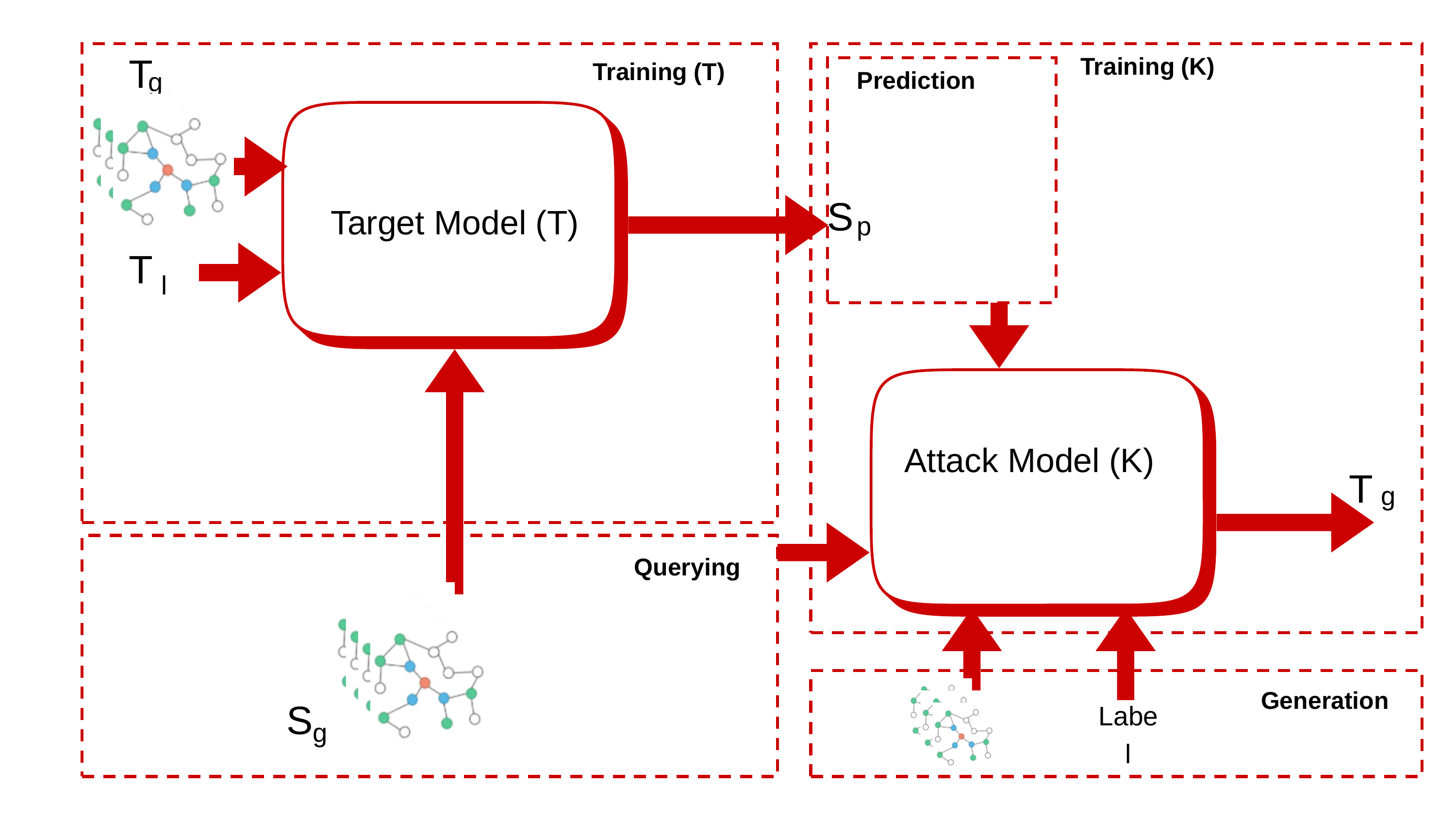}
        \label{fig:sub1}
        \caption{Graph \& Label-Conditioned (GLC) Attack}
    \end{subfigure}
    \begin{subfigure}[t]{0.88\columnwidth}
        \centering
        \includegraphics[width=\linewidth]{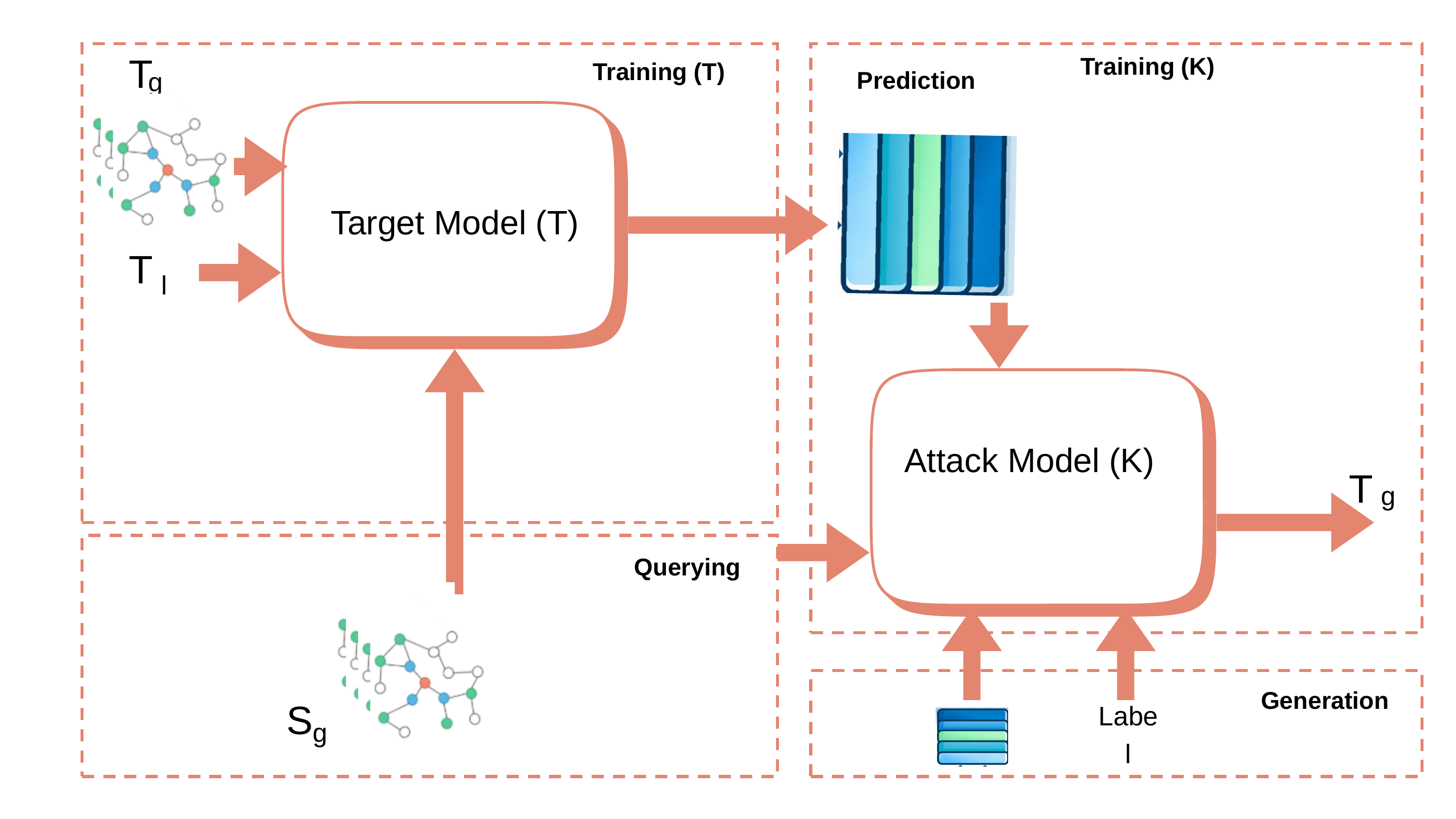}
        \label{fig:sub2}
        \caption{Embedding \& Label-Conditioned (ELC) Attack}
    \end{subfigure}
    
    \caption{An Overview of the proposed two novel privacy attacks against graphs: (a) GLC attack and (b) ELC attack.}
    \label{fig:framework}
\end{figure*}
=================}

\subsection{\textcolor{black}{Threat Model}}\label{subsec:t_model}
\subsubsection{Adversary Goal}
The graph data for training the target model are sensitive and private. The adversary aims to reconstruct the training graph data of the target model using predictions from the model and graph embeddings. 

\subsubsection{Adversary Capability and Assumption}
\label{subsubsec:advers_capab}
We consider a model inversion attack under black-box settings. We assumed that the adversary could query the target model and obtain the prediction and intermediate representation to generate graph data embeddings. We also assume that the adversary has access to in-distribution graph data.\\

We propose two novel graph reconstruction attacks: \textcolor{black}{graph-label conditioned (GLC)} attack and \textcolor{black}{embedding-label conditioned (ELC)} attack to reconstruct GNNs ($Q_m^{\phi}$) training graph (data). However, we take into account \textcolor{black}{scenarios} in which the adversary has access to the in-distribution data and is limited to querying the victim model; a setting commonly referred to as \textit{black-box}. By merely querying the target model to predict the data class, which is utilized in conjunction with the query graph to train the attack model conditioned on the graph and label. \textcolor{black}{The \textcolor{black}{GLC} attack is intended to recover sensitive data information from the target model. Also, querying the target model to output the graph embeddings, the \textcolor{black}{ELC} is intended to reconstruct the target model's sensitive training graph data.} 
\subsection{Attack Overview} \label{subsec:attack_overview}
\subsubsection{Graph-label conditioned}
Here, the reconstruction of the target model training graph data is represented as an adversarial game between the discriminator and the generator while leveraging the structural information and predicted labels from a target model to recover sensitive graph data. \textcolor{black}{Figure~\ref{fig:sub1}} illustrates the \textcolor{black}{GLC} attack framework.\\
\textbf{Model Definition}
Let $\mathcal{G} = (\mathbf{A}, \mathbf{X})$ represent a graph, where $\mathbf{X} \in \mathbb{R}^{N \times d}$ represents node features, $\mathbf{A} \in \{0,1\}^{N \times N}$ is the adjacency matrix and \textcolor{black}{$\phi$ represent the model parameter}. The target model can be defined as:
\begin{equation}
Q_m^{\phi} : \mathcal{G} \rightarrow \mathcal{Y}
\quad
y = Q_m^{\phi}(\mathcal{G})
\end{equation}

Given an in-distribution dataset:
\begin{equation}
 \textcolor{black}{\mathcal{D} = \{(\mathcal{G}_i, \mathcal{Y}_i)\}_{i=1}^{N} } 
\end{equation}
The adversary obtains the predicted labels by querying the target model:
\begin{equation}
 \textcolor{black}{y_i = Q_m^{\phi}(\mathcal{G}_i)}
\end{equation}
 \textcolor{black}{The resulting predicted label is used together with its graph to form the attack dataset:}
\begin{equation}
\mathcal{D}_{\text{attack}} = \{(\mathcal{G}_i, y_i)\}_{i=1}^{N}
\end{equation}

\textcolor{black}{The \textcolor{black}{GLC} attack is modeled using a generator S$_{\phi}$ and discriminator $K_{\psi}$.}\\
\textbf{Generator:} The generator reconstructs a graph conditioned on the input graph, predicted label, and noise:
\begin{equation}
\textcolor{black}{\hat{\mathcal{G}} = S_{\phi}(\mathcal{G}, y, z), \quad z \sim p_z(z)}
\end{equation}
\textcolor{black}{where $\hat{\mathcal{G}}$ represent the reconstructed graph.}\\
\textbf{Discriminator:}
The discriminator distinguishes between real and generated graph-label pairs:
\begin{equation}
\textcolor{black}{K_{\psi}(\mathcal{G}, \mathcal{Y}) \in [0,1], \quad K_{\psi}(\hat{\mathcal{G}}, y) \in [0,1]}
\end{equation}
\textbf{Adversarial Objective:}
The GAN is trained using the minimax objective:
\textcolor{black}{\begin{equation}
\begin{aligned}
\min_{\phi} \max_{\psi} \; \mathcal{L}_{\text{adv}} =
&\ \mathbb{E}_{(\mathcal{G}, \mathcal{Y}) \sim \mathcal{D}_{\text{attack}}}
\left[ \log K_{\psi}(\mathcal{G}, \mathcal{Y}) \right] \\
&+ \mathbb{E}_{z \sim p_z,\; \mathcal{G} \sim \mathcal{D}_{\text{attack}}} \Big[
\log \big(1 - K_{\psi}(\hat{\mathcal{G}}, y)\big)
\Big]
\end{aligned}
\end{equation}}
\textcolor{black}{where $\mathcal{L}_{\text{adv}}$ is the adversarial loss}.
Optimizing the generator using non-saturated loss:
\textcolor{black}{\begin{equation}
\mathcal{L}_G =
\mathbb{E}_{z \sim p_z}
\left[ -\log Q_m^{\phi}(K_{\psi}(\mathcal{K}, y, z), y) \right]
\end{equation}}\\
\textcolor{black}{where $\mathcal{K}$ represents leaked information available to the adversary, $y$ represents the predicted label, $Q_m^\phi$ represents the target model with it parameter and $\mathcal{L}_G$ is the generator loss.} \\
\textbf{Optimization Goal:}
Solving ($\min_{\phi} \max_{\psi} \; \mathcal{L}_{\text{adv}}$) will enable the adversary to recover the underlying graph distribution. At convergence, the generator approximates the conditional distribution, enabling reconstruction of sensitive structural properties of the original training graph
\textcolor{black}{$p(\hat{\mathcal{G}} \mid y$)}.

\subsubsection{Embedding-label conditioned}
Given the intermediate representations derived from in-distribution data of the target model, the \textcolor{black}{ELC} attack reconstructs sensitive private graph data using both graph embeddings and their labels. \textcolor{black}{Figure~\ref{fig:sub2}} illustrates the overview of the \textcolor{black}{ELC} attack.\\
\textbf{Model Definition:}
Let $\mathcal{G} = (\mathbf{A}, \mathbf{X})$ represent a graph, where $\mathbf{X} \in \mathbb{R}^{N \times d}$ represents node features and $\mathbf{A} \in \{0,1\}^{N \times N}$ is the adjacency matrix. The target model can be defined as:
\begin{equation}
 \textcolor{black}{Q_m^{\phi} : \mathcal{G} \rightarrow \mathcal{H} }
\end{equation}
where $\mathbf{h} \in \mathcal{H}$ denotes the intermediate embedding. 
Given an in-distribution dataset:
\begin{equation}
\textcolor{black}{\mathcal{D} = \{(\mathcal{G}_i, \mathcal{Y}_i)\}_{i=1}^{N}}
\end{equation}
The adversary obtains the embeddings by querying the
target model:

\begin{equation}
\textcolor{black}{h_i = Q_m^{\phi}(\mathcal{G}_i)}
\end{equation}
Thus, the attack dataset is constructed as:
\begin{equation}
\mathcal{D}_{\text{attack}} = \{(\mathcal{H}_i \mathcal{Y}_i)\}_{i=1}^{N}
\end{equation}

\ignore{
\label{sec:exp}
In this section, we discuss our setup and the dataset used for the study. We highlight and define four (4) metrics to assess our novel attack methods on the NCI1 dataset. 
\begin{figure*}[t]
    \centering
    \includegraphics[width=0.9\textwidth]{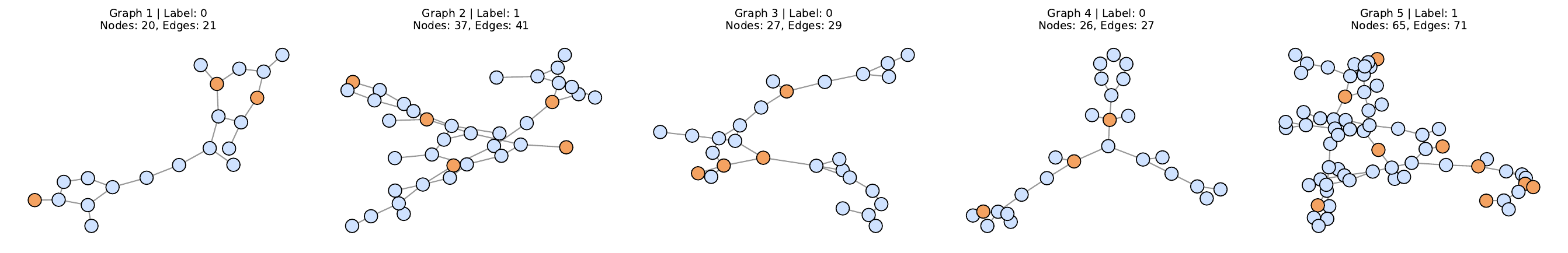}

    \vspace{0.6em}

    \includegraphics[width=0.9\textwidth]{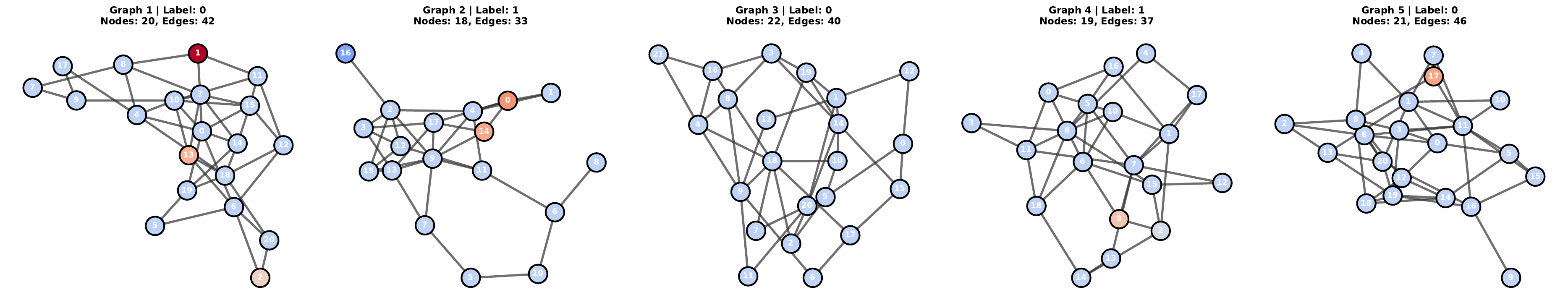}

    \vspace{0.6em}

    \includegraphics[width=0.9\textwidth]{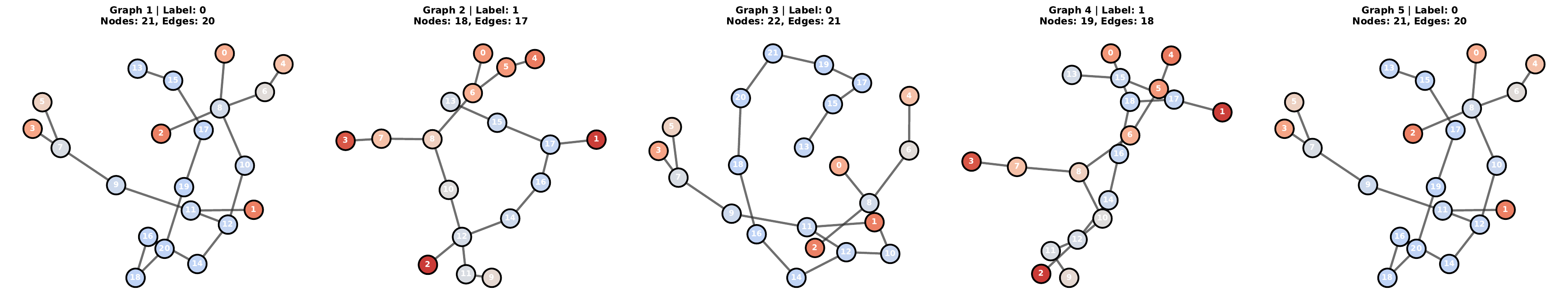}

    \caption{Sample of the reconstructed graph.}
    \label{fig:glgra_different_double}
\end{figure*}
}

The \textcolor{black}{ELC} attack is modeled using a generator $S_{\phi}$ and discriminator $K_{\psi}$.\\
\textcolor{black}{\textbf{Generator:} The generator reconstructs the training graph when conditioned on noise, labels, and embeddings:}
\begin{equation}
\hat{\mathcal{G}} = S_{\phi}(\mathbf{h}, y, z), \quad z \sim p_z(z)
\end{equation}
\textbf{Discriminator:} The discriminator distinguishes between real and reconstructed graph-embedding pairs:
\begin{equation}
\textcolor{black}{K_{\psi}(\mathcal{H}, \mathcal{Y}) \in [0,1], \quad K_{\psi}(\hat{\mathcal{H}}, \mathcal{Y}) \in [0,1]}
\end{equation}
\textbf{Adversarial Objective:}
The GAN is trained using the minimax objective:
\begin{equation}
\begin{aligned}
\min_{\phi}\max_{\psi}\;
\mathcal{L}_{\mathrm{adv}}
=&\;
\mathbb{E}
_{\substack{(\mathcal{G},y)\sim\mathcal{D}_{\mathrm{attack}}}}
\!\left[
\log K_{\psi}(\mathcal{G},y)
\right]
\\
&+
\mathbb{E}
_{\substack{(\mathbf{h},y)\sim\mathcal{D}_{\mathrm{attack}}\\
z\sim p_z}}
\!\left[
\log\!\left(
1-K_{\psi}(\hat{\mathcal{G}},y)
\right)
\right]
\end{aligned}
\end{equation}
\textcolor{black}{Therefore, we optimize the generator using:}
\begin{equation}
\mathcal{L}_{S} =
\mathbb{E}_{\mathbf{h}, y, z}
\left[ -\log K_{\psi}(S_{\phi}(\mathbf{h}, y, z), y) \right]
\end{equation}
\textcolor{black}{where $\mathcal{L}_{S}$ represents label consistency loss.\\
\textbf{Reconstruction Objective: }}
A reconstruction loss is introduced:
\begin{equation}
\mathcal{L}_{\text{rec}} =
\mathbb{E}\left[ \mathcal{L}\big(\mathcal{G}, S_{\phi}(\mathbf{h}, y, z)\big) \right]
\end{equation}
\textcolor{black}{\textbf{Final Objective}
optimization becomes:
\begin{equation}
\min_{\phi} \max_{\psi} \; \mathcal{L}_{\text{S}} + \lambda \mathcal{L}_{\text{rec}}
\end{equation}
where $\lambda$ is a trade-off hyperparameter that balances the label consistency objective and the graph reconstruction objective.
The reconstructed graph is defined as:
\begin{equation}
S_{\phi} : \mathcal{H} \times \mathcal{Y} \times \mathcal{Z}
\rightarrow \mathcal{G}, \hat{\mathcal{G}} = S_{\phi}(\mathbf{h}, y, z)
\end{equation}
}

\begin{table}[b]
\caption{Summary of the datasets used in the experiments.}
\centering
\renewcommand{\arraystretch}{1.1}
\setlength{\tabcolsep}{3pt}
\begin{tabular}{lcccc}
\hline
\textbf{Dataset} & \textbf{\#Graphs} & \textbf{Nodes} & \textbf{Edges} & \textbf{Classes} \\
\hline
NCI1     & 4,110 & 29.87 & 32.30 & 2 \\
AIDS     & 2,000 & 15.69 & 32.39 & 2 \\
PROTEINS & 1,113 & 39.06 & 72.82 & 2 \\
\hline
\end{tabular}
\label{tab:datasets}
\end{table}

\section{Experiments}
\vspace{-0.5cm}

\textcolor{black}{\subsection{Experimental Setup}
\label{subsec:exp_setup}
We used the PyTorch framework for implementation in this study. The implementation was performed on a high-performance computing (HPC) system with an NVIDIA A100 Tensor Core GPU and 60GB of RAM. The model was developed and assessed using the NCI1, AIDS, and PROTEINS datasets. This drug discovery research dataset is associated with a class that could have biological or dangerous effects. The datasets were divided into target and shadow sets at a 1:1 ratio. The attack model was trained and tested using the shadow dataset, whereas the target model was trained and validated using the target dataset. The adversarial goal is to apply the model inversion attack approach to infer the training data of the target GNN model.} 

\subsection{Datasets}\label{subsec:datasets_all}
Three datasets were used to carry out the empirical analysis in this study. The NCI1 is a molecular graph where nodes represent atoms and edges denote chemical bonds. AIDS is also a molecular graph from the AIDS Antiviral Screen database. Nodes represent atoms and edges represent chemical bonds. Graphs are classified as active or inactive against HIV. The PROTEINS data nodes represent secondary structural elements and are named according to their kind, which can be sheet, turn, or helix. A graph edge joins two nodes if they are among the three nearest neighbors in space or in the amino acid sequence. \textcolor{black}{Table~\ref{tab:datasets}} summarizes the dataset used in this study.

\begin{figure*}[!tbh]
\centering
\begin{subfigure}{0.23\textwidth}
    \centering
    \includegraphics[width=\linewidth]{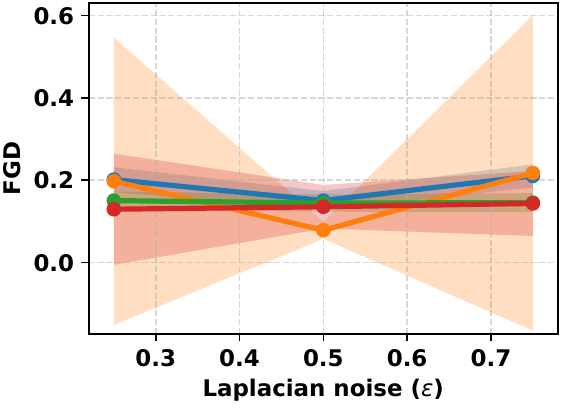}
 
\end{subfigure}
\hfill
\begin{subfigure}{0.23\textwidth}
    \centering
    \includegraphics[width=\linewidth]{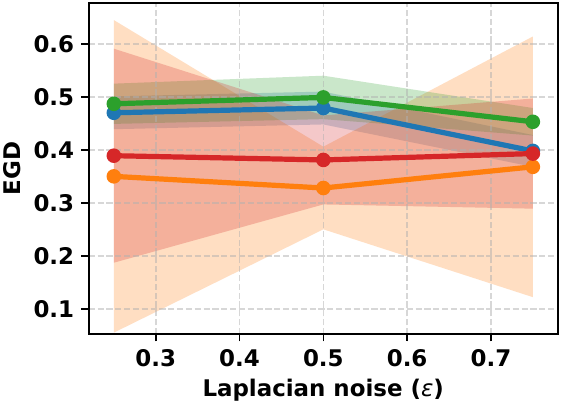}
    
\end{subfigure}
\hfill
\begin{subfigure}{0.23\textwidth}
    \centering
    \includegraphics[width=\linewidth]{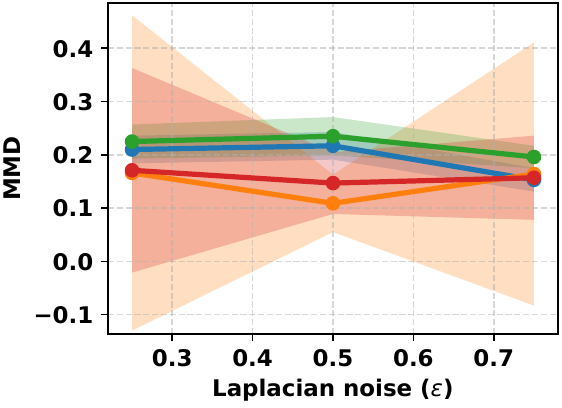}
  
\end{subfigure}
\hfill
\begin{subfigure}{0.23\textwidth}
    \centering
    \includegraphics[width=\linewidth]{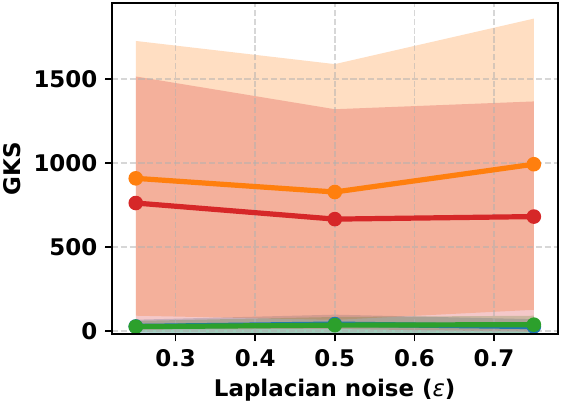}
  
\end{subfigure}

\vspace{2mm}

\textbf{(a) NCI1}

\vspace{3mm}

\begin{subfigure}{0.23\textwidth}
    \centering
    \includegraphics[width=\linewidth]{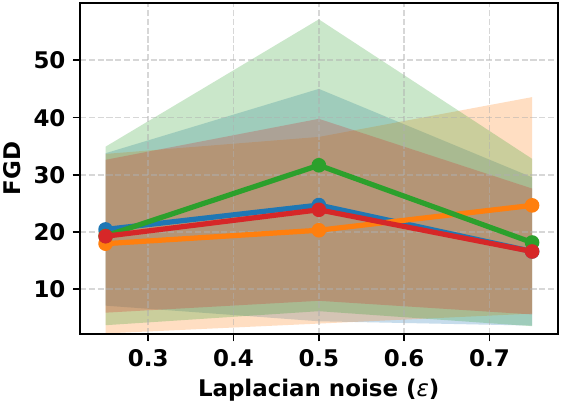}
\end{subfigure}
\hfill
\begin{subfigure}{0.23\textwidth}
    \centering
    \includegraphics[width=\linewidth]{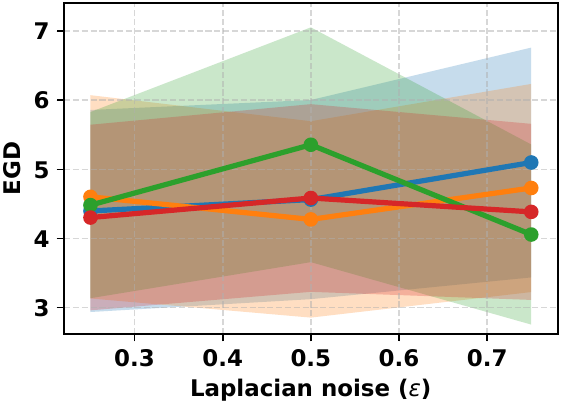}
\end{subfigure}
\hfill
\begin{subfigure}{0.23\textwidth}
    \centering
    \includegraphics[width=\linewidth]{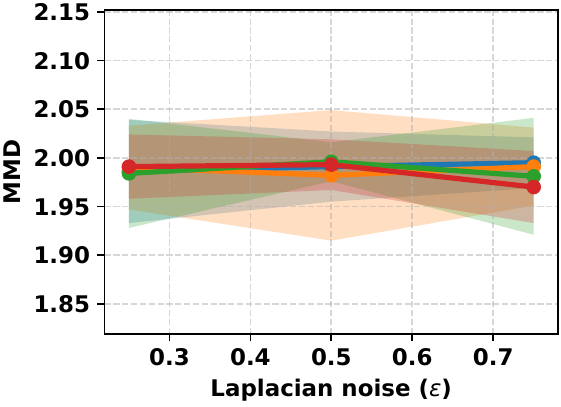}
\end{subfigure}
\hfill
\begin{subfigure}{0.23\textwidth}
    \centering
    \includegraphics[width=\linewidth]{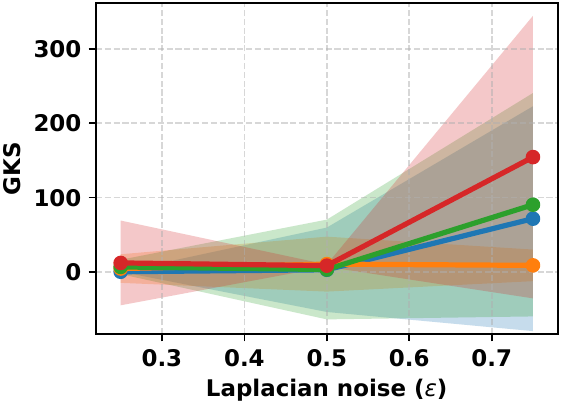}
\end{subfigure}

\vspace{2mm}

\textbf{(b) PROTEINS}

\vspace{3mm}

\begin{subfigure}{0.23\textwidth}
    \centering
    \includegraphics[width=\linewidth]{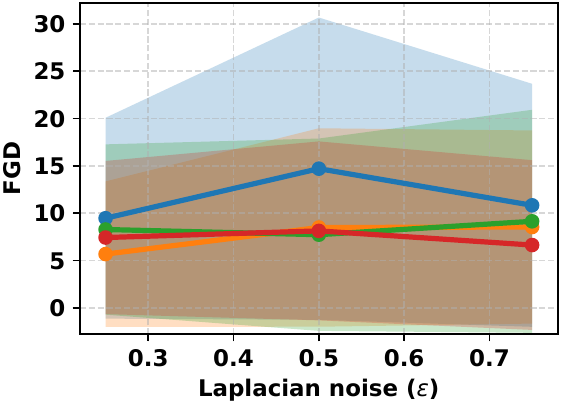}
\end{subfigure}
\hfill
\begin{subfigure}{0.23\textwidth}
    \centering
    \includegraphics[width=\linewidth]{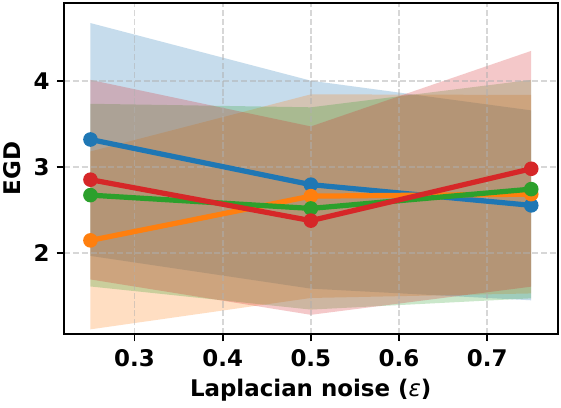}
\end{subfigure}
\hfill
\begin{subfigure}{0.23\textwidth}
    \centering
    \includegraphics[width=\linewidth]{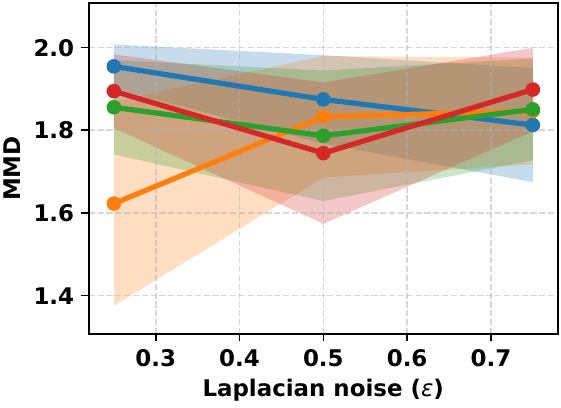}
\end{subfigure}
\hfill
\begin{subfigure}{0.23\textwidth}
    \centering
    \includegraphics[width=\linewidth]{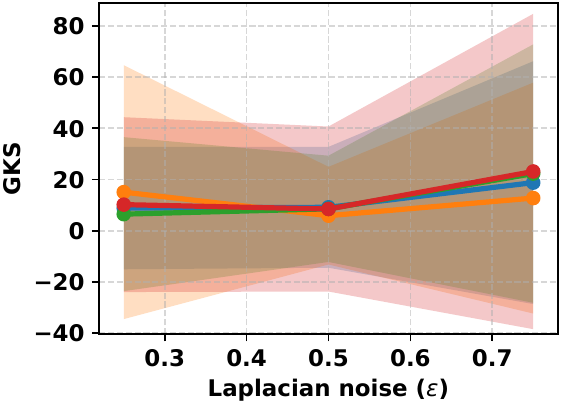}
\end{subfigure}

\textbf{(c) AIDS}
\begin{center}
    \includegraphics[width=0.5\textwidth]{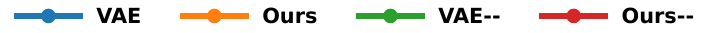}
\end{center}
\caption{\textcolor{black}{Comparison of graph reconstruction quality under \textcolor{black}{GLC} attack with varying Laplace scales for three datasets. Columns correspond to FGD, EGD, MMD, and GKS, respectively. Rows correspond to NCI1, PROTEINS, and AIDS datasets.}}
\label{fig:GLC_plot}
\end{figure*}

\begin{table*}[ht!]
\caption{\textcolor{black}{Comparative results of Ours \& Ours$--$ and baseline attacks on NCI1, PROTEINS, and AIDS datasets with varying \textit{Laplacian} noise ($\epsilon$) value for both \textcolor{black}{GLC} and \textcolor{black}{ELC} attacks}. The \colorbox{paretoyellow}{yellow} color represents the best-performing technique for each metric within each attack type.}
\centering
\resizebox{\textwidth}{!}{%
\begin{tabular}{|c|c|c|cc|cc|cc|cc|}
\hline
\multirow{2}{*}{Method} & \multirow{2}{*}{Dataset} & \multirow{2}{*}{$\epsilon$}
& \multicolumn{2}{c|}{FGD $\downdownarrows$} & \multicolumn{2}{c|}{EGD $\downdownarrows$} & \multicolumn{2}{c|}{MMD $\downdownarrows$} & \multicolumn{2}{c|}{GKS $\upuparrows$}\\
\cline{4-11}
& & & GLC & ELC & GLC & ELC & GLC & ELC & GLC & ELC \\
\hline


VAE~~~~~ & \multirow{12}{*}{NCI1} & 0.25
& 0.201 $\pm$ 0.030 & 0.254 $\pm$ 0.041
& 0.470 $\pm$ 0.031 & 0.494 $\pm$ 0.042
& 0.210 $\pm$ 0.026 & 0.230 $\pm$ 0.036
& 28.497 $\pm$ 40.111 & 0.60 $\pm$ 0.692\\

VAE$--$ && 0.25
& 0.150 $\pm$ 0.022 & 0.237 $\pm$ 0.034
& 0.487 $\pm$ 0.038 & 0.423 $\pm$ 0.042
& 0.225 $\pm$ 0.032 & 0.173 $\pm$ 0.032
& 27.103 $\pm$ 34.319 & 0.047 $\pm$ 0.997\\

Ours~~~~ && 0.25
& 0.197 $\pm$ 0.349 & 0.227 $\pm$ 0.077
& \cellcolor{paretoyellow}\textbf{\makecell{0.350 $\pm$ 0.295}} & 0.451 $\pm$ 0.064
& \cellcolor{paretoyellow}\textbf{\makecell{0.166 $\pm$ 0.296}} & 0.197 $\pm$ 0.052
& \cellcolor{paretoyellow}\textbf{\makecell{910.307 $\pm$ 817.914}} & \cellcolor{paretoyellow}\textbf{\makecell{31.649 $\pm$ 24.388}}\\

Ours$--$ && 0.25
& \cellcolor{paretoyellow}\textbf{\makecell{0.129 $\pm$ 0.135}} & \cellcolor{paretoyellow}\textbf{\makecell{0.223 $\pm$ 0.084}}
& 0.389 $\pm$ 0.202 & \cellcolor{paretoyellow}\textbf{\makecell{0.395 $\pm$ 0.078}}
& 0.171 $\pm$ 0.192 & \cellcolor{paretoyellow}\textbf{\makecell{0.155 $\pm$ 0.053}}
& 763.318 $\pm$ 753.482 & 28.88 $\pm$ 26.46\\
\cline{1-1}\cline{3-11}

VAE~~~~~ && 0.50
& 0.150 $\pm$ 0.023 & 0.246 $\pm$ 0.041
& 0.479 $\pm$ 0.031 & \cellcolor{paretoyellow}\textbf{\makecell{0.381 $\pm$ 0.031}}
& 0.217 $\pm$ 0.026 & \cellcolor{paretoyellow}\textbf{\makecell{0.141 $\pm$ 0.021}}
& 42.419 $\pm$ 56.160 & 0.02 $\pm$ 0.692\\

VAE$--$ && 0.50
& 0.144 $\pm$ 0.021 & 0.184 $\pm$ 0.026
& 0.499 $\pm$ 0.041 & 0.460 $\pm$ 0.043
& 0.235 $\pm$ 0.036 & 0.202 $\pm$ 0.035
& 36.154 $\pm$ 48.610 & 0.082 $\pm$ 1.929\\

Ours~~~~ && 0.50
& \cellcolor{paretoyellow}\textbf{\makecell{0.078 $\pm$ 0.021}} & 0.179 $\pm$ 0.051
& \cellcolor{paretoyellow}\textbf{\makecell{0.328 $\pm$ 0.078}} & 0.383 $\pm$ 0.041
& \cellcolor{paretoyellow}\textbf{\makecell{0.109 $\pm$ 0.054}} & 0.143 $\pm$ 0.029
& \cellcolor{paretoyellow}\textbf{\makecell{828.920 $\pm$ 761.401}} & \cellcolor{paretoyellow}\textbf{\makecell{27.970 $\pm$ 40.093}}\\

Ours$--$ && 0.50
& 0.135 $\pm$ 0.053 & \cellcolor{paretoyellow}\textbf{\makecell{0.178 $\pm$ 0.039}}
& 0.381 $\pm$ 0.084 & 0.439 $\pm$ 0.060
& 0.147 $\pm$ 0.058 & 0.186 $\pm$ 0.051
& 667.487 $\pm$ 654.333 & 7.670 $\pm$ 9.249\\
\cline{1-1}\cline{3-11}

VAE~~~~~ && 0.75
& 0.210 $\pm$ 0.028 & 0.246 $\pm$ 0.041
& 0.398 $\pm$ 0.030 & 0.456 $\pm$ 0.034
& \cellcolor{paretoyellow}\textbf{\makecell{0.153 $\pm$ 0.022}} & 0.198 $\pm$ 0.027
& 27.730 $\pm$ 43.919 & 0.20 $\pm$ 0.692\\

VAE$--$ && 0.75
& 0.145 $\pm$ 0.023 & 0.214 $\pm$ 0.039
& 0.453 $\pm$ 0.026 & 0.490 $\pm$ 0.045
& 0.196 $\pm$ 0.021 & 0.227 $\pm$ 0.039
& 39.122 $\pm$ 53.124 & 0.082 $\pm$ 1.939\\

Ours~~~~ && 0.75
& 0.217 $\pm$ 0.383 & 0.223 $\pm$ 0.060
& \cellcolor{paretoyellow}\textbf{\makecell{0.368 $\pm$ 0.246}} & \cellcolor{paretoyellow}\textbf{\makecell{0.435 $\pm$ 0.034}}
& 0.164 $\pm$ 0.247 & \cellcolor{paretoyellow}\textbf{\makecell{0.182 $\pm$ 0.027}}
& \cellcolor{paretoyellow}\textbf{\makecell{994.390 $\pm$ 866.768}} & 16.077 $\pm$ 18.818\\

Ours$--$ && 0.75
& \cellcolor{paretoyellow}\textbf{\makecell{0.143 $\pm$ 0.079}} & \cellcolor{paretoyellow}\textbf{\makecell{0.198 $\pm$ 0.053}}
& 0.393 $\pm$ 0.104 & 0.461 $\pm$ 0.032
& 0.157 $\pm$ 0.079 & 0.202 $\pm$ 0.027
& 682.626 $\pm$ 685.475 & \cellcolor{paretoyellow}\textbf{\makecell{17.216 $\pm$ 10.899}}\\
\hline

VAE~~~~~ & \multirow{12}{*}{AIDS} & 0.25
& 9.466 $\pm$ 10.611 & 9.509 $\pm$ 10.782
& 3.320 $\pm$ 1.356 & 2.838 $\pm$ 1.206
& 1.954 $\pm$ 0.053 & 1.884 $\pm$ 0.101
& 8.844 $\pm$ 23.904 & \cellcolor{paretoyellow}\textbf{\makecell{303.815 $\pm$ 326.927}}\\

VAE$--$ && 0.25
& 8.272 $\pm$ 8.996 & 14.045 $\pm$ 10.231
& 2.672 $\pm$ 1.064 & 3.133 $\pm$ 1.322
& 1.855 $\pm$ 0.114 & 1.929 $\pm$ 0.342
& 6.477 $\pm$ 30.072 & 0.000 $\pm$ 0.000\\

Ours~~~~ && 0.25
& \cellcolor{paretoyellow}\textbf{\makecell{5.675 $\pm$ 7.682}} & 6.975 $\pm$ 7.835
& \cellcolor{paretoyellow}\textbf{\makecell{2.145 $\pm$ 1.036}} & 2.436 $\pm$ 1.020
& \cellcolor{paretoyellow}\textbf{\makecell{1.622 $\pm$ 0.247}} & 1.778 $\pm$ 0.162
& \cellcolor{paretoyellow}\textbf{\makecell{15.071 $\pm$ 49.590}} & 0.040 $\pm$ 0.434\\

Ours$--$ && 0.25
& 7.424 $\pm$ 8.091 & \cellcolor{paretoyellow}\textbf{\makecell{6.621 $\pm$ 8.114}}
& 2.852 $\pm$ 1.162 & \cellcolor{paretoyellow}\textbf{\makecell{2.344 $\pm$ 1.062}}
& 1.894 $\pm$ 0.089 & \cellcolor{paretoyellow}\textbf{\makecell{1.728 $\pm$ 0.187}}
& 10.191 $\pm$ 34.150 & 0.980 $\pm$ 2.812\\
\cline{1-1}\cline{3-11}

VAE~~~~~ && 0.50
& 14.698 $\pm$ 15.970 & \cellcolor{paretoyellow}\textbf{\makecell{7.409 $\pm$ 7.369}}
& 2.794 $\pm$ 1.211 & 2.541 $\pm$ 0.976
& 1.874 $\pm$ 0.107 & 1.823 $\pm$ 0.130
& \cellcolor{paretoyellow}\textbf{\makecell{9.148 $\pm$ 23.609}} & 0.000 $\pm$ 0.000\\

VAE$--$ && 0.50
& \cellcolor{paretoyellow}\textbf{\makecell{7.723 $\pm$ 10.161}} & 11.490 $\pm$ 9.231
& 2.517 $\pm$ 1.178 & \cellcolor{paretoyellow}\textbf{\makecell{2.425 $\pm$ 1.324}}
& 1.786 $\pm$ 0.158 & \cellcolor{paretoyellow}\textbf{\makecell{1.759 $\pm$ 1.022}}
& 8.550 $\pm$ 20.755 & 0.000 $\pm$ 0.000\\

Ours~~~~ && 0.50
& 8.483 $\pm$ 10.483 & 8.066 $\pm$ 9.459
& 2.660 $\pm$ 1.186 & 2.603 $\pm$ 1.135
& 1.832 $\pm$ 0.147 & 1.824 $\pm$ 0.143
& 5.882 $\pm$ 19.074 & 4.700 $\pm$ 5.211\\

Ours$--$ && 0.50
& 8.125 $\pm$ 9.453 & 8.115 $\pm$ 9.242
& \cellcolor{paretoyellow}\textbf{\makecell{2.375 $\pm$ 1.097}} & 2.622 $\pm$ 1.114
& \cellcolor{paretoyellow}\textbf{\makecell{1.744 $\pm$ 0.171}} & 1.831 $\pm$ 0.144
& 8.478 $\pm$ 32.227 & \cellcolor{paretoyellow}\textbf{\makecell{13.455 $\pm$ 30.461}}\\
\cline{1-1}\cline{3-11}

VAE~~~~~ && 0.75
& 10.828 $\pm$ 12.838 & 8.772 $\pm$ 8.020
& \cellcolor{paretoyellow}\textbf{\makecell{2.553 $\pm$ 1.106}} & 2.784 $\pm$ 1.011
& \cellcolor{paretoyellow}\textbf{\makecell{1.812 $\pm$ 0.138}} & 1.886 $\pm$ 0.100
& 18.782 $\pm$ 47.426 & 0.000 $\pm$ 0.000\\

VAE$--$ && 0.75
& 9.138 $\pm$ 11.782 & 11.358 $\pm$ 9.012
& 2.742 $\pm$ 1.272 & 2.875 $\pm$ 1.231
& 1.850 $\pm$ 0.123 & 1.882 $\pm$ 0.987
& 22.316 $\pm$ 50.474 & 0.000 $\pm$ 0.000\\

Ours~~~~ && 0.75
& 8.535 $\pm$ 10.203 & 4.227 $\pm$ 4.368
& 2.683 $\pm$ 1.156 & 1.897 $\pm$ 0.793
& 1.846 $\pm$ 0.129 & 1.502 $\pm$ 0.311
& 12.751 $\pm$ 45.104 & \cellcolor{paretoyellow}\textbf{\makecell{12.750 $\pm$ 14.074}}\\

Ours$--$ && 0.75
& \cellcolor{paretoyellow}\textbf{\makecell{6.630 $\pm$ 8.967}} & \cellcolor{paretoyellow}\textbf{\makecell{3.858 $\pm$ 4.622}}
& 2.979 $\pm$ 1.373 & \cellcolor{paretoyellow}\textbf{\makecell{1.778 $\pm$ 0.834}}
& 1.898 $\pm$ 0.101 & \cellcolor{paretoyellow}\textbf{\makecell{1.401 $\pm$ 0.323}}
& \cellcolor{paretoyellow}\textbf{\makecell{23.132 $\pm$ 61.584}} & 2.185 $\pm$ 3.502\\
\hline

VAE~~~~~ & \multirow{12}{*}{PROTEINS} & 0.25
& 20.438 $\pm$ 13.309 & 28.807 $\pm$ 31.110
& 4.397 $\pm$ 1.463 & 5.016 $\pm$ 1.910
& 1.986 $\pm$ 0.053 & 1.995$\pm$ 0.014
& 0.000 $\pm$ 0.0000 & 20.815 $\pm$ 70.927\\

VAE$--$ && 0.25
& 19.302 $\pm$ 15.587 & 21.400$\pm$ 15.232
& 4.482 $\pm$ 1.348 & 4.090$\pm$ 1.764
&  \cellcolor{paretoyellow}\textbf{\makecell{1.984 $\pm$ 0.056}} & 1.979 $\pm$ 0.678
& 6.487 $\pm$ 9.654 & 0.000 $\pm$ 0.000\\

Ours~~~~ && 0.25
& \cellcolor{paretoyellow}\textbf{\makecell{17.951 $\pm$ 15.674}} & \cellcolor{paretoyellow}\textbf{\makecell{18.881 $\pm$ 21.201}}
& 4.602 $\pm$ 1.472 & \cellcolor{paretoyellow}\textbf{\makecell{3.994$\pm$ 1.711}}
&1.990 $\pm$ 0.043 &  \cellcolor{paretoyellow}\textbf{\makecell{1.974 $\pm$ 0.052}}
&4.309 $\pm$ 19.226 &  \cellcolor{paretoyellow}\textbf{\makecell{190.161 $\pm$ 157.538}}\\

Ours$--$ && 0.25
& 19.249 $\pm$ 13.346 & 19.079 $\pm$ 21.227
& \cellcolor{paretoyellow}\textbf{\makecell{4.302 $\pm$ 1.344}} & 4.020 $\pm$ 1.709
& 1.991 $\pm$ 0.033& 1.976 $\pm$ 0.049
&  \cellcolor{paretoyellow}\textbf{\makecell{11.877$\pm$ 56.985}} & 171.23$\pm$ 119.8\\
\cline{1-1}\cline{3-11}

VAE~~~~~ && 0.50
& 24.691 $\pm$ 20.292 & \cellcolor{paretoyellow}\textbf{\makecell{20.845$\pm$ 21.109}}
& 4.562 $\pm$ 1.442 & 4.795 $\pm$ 1.547
& 1.991 $\pm$ 0.036 & 1.990 $\pm$ 0.026
& 2.671 $\pm$ 56.765 & 0.000 $\pm$ 0.000\\

VAE$--$ && 0.50
&31.643 $\pm$ 25.481 & 19.509 $\pm$ 15.345
& 5.356 $\pm$ 1.70 &4.609$\pm$ 2.542
& 1.996 $\pm$  0.020 & 1.991 $\pm$ 0.633
& 3.029 $\pm$ 67.085 & 0.022 $\pm$ 0.001\\

Ours~~~~ && 0.50
&  \cellcolor{paretoyellow}\textbf{\makecell{20.295 $\pm$ 16.298 }}& 25.159 $\pm$ 28.412
&  \cellcolor{paretoyellow}\textbf{\makecell{4.275 $\pm$ 1.422 }}& 4.605 $\pm$ 1.987
&  \cellcolor{paretoyellow}\textbf{\makecell{1.982 $\pm$ 0.067}} & 1.989 $\pm$ 0.028
&  \cellcolor{paretoyellow}\textbf{\makecell{10.264 $\pm$ 36.966}} & 98.952$\pm$ 148.572\\

Ours$--$ && 0.50
& 23.866$\pm$ 15.889 & 25.053 $\pm$ 28.430
& 4.585 $\pm$ 1.359 &  \cellcolor{paretoyellow}\textbf{\makecell{4.592 $\pm$ 1.992}}
& 1.993 $\pm$ 0.026 &  \cellcolor{paretoyellow}\textbf{\makecell{1.988$\pm$ 0.031}}
& 8.191 $\pm$ 1.549 & \cellcolor{paretoyellow}\textbf{\makecell{99.010 $\pm$ 107.490}}\\
\cline{1-1}\cline{3-11}

VAE~~~~~ && 0.75
& 16.595 $\pm$ 12.946 & 28.010 $\pm$ 29.095
& 5.099 $\pm$ 1.664 & 4.968 $\pm$ 1.824
& 1.995 $\pm$ 0.026 & 1.996 $\pm$ 0.011
& 71.375 $\pm$ 151.200.4 & 0.000 $\pm$ 0.000\\

VAE$--$ && 0.75
& 18.155 $\pm$ 14.612 & 22.258 $\pm$ 19.021
&  \cellcolor{paretoyellow}\textbf{\makecell{4.056 $\pm$ 1.304}} &  3.862 $\pm$ 2.011        &  1.981 $\pm$ 0.060 & 1.974 $\pm$ 0.043
& 90.151 $\pm$ 150.221 &96.907$\pm$ 176.994 \\

Ours~~~~ && 0.75
& 24.638 $\pm$ 18.905&  \cellcolor{paretoyellow}\textbf{\makecell{15.534 $\pm$ 16.632}}
& 4.730 $\pm$ 1.507 &  \cellcolor{paretoyellow}\textbf{\makecell{3.642 $\pm$ 1.505}}
& 1.991 $\pm$ 0.040 &  \cellcolor{paretoyellow}\textbf{\makecell{1.957 $\pm$ 0.076}}
& 8.601 $\pm$ 21.257 & 144.429 $\pm$ 122.340\\

Ours$--$ && 0.75
& \cellcolor{paretoyellow}\textbf{\makecell{16.587 $\pm$ 10.992}} & 22.677 $\pm$ 17.256
& 4.384 $\pm$ 1.275 & 4.516 $\pm$ 1.512
&  \cellcolor{paretoyellow}\textbf{\makecell{1.970 $\pm$ 0.037 }}& 1.987 $\pm$ 0.044
&  \cellcolor{paretoyellow}\textbf{\makecell{154.212 $\pm$ 190 }}&  \cellcolor{paretoyellow}\textbf{\makecell{184.625$\pm$ 237.854}}\\
\hline
\end{tabular}%
}
\label{table:merged}
\end{table*}

\begin{figure*}[t]
\centering
\begin{subfigure}{0.23\textwidth}
    \centering
    \includegraphics[width=\linewidth]{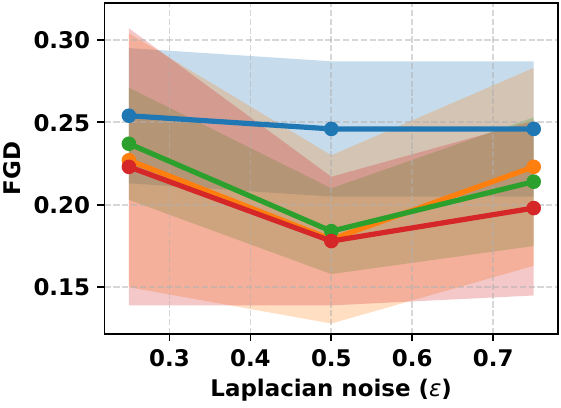}
 
\end{subfigure}
\hfill
\begin{subfigure}{0.23\textwidth}
    \centering
    \includegraphics[width=\linewidth]{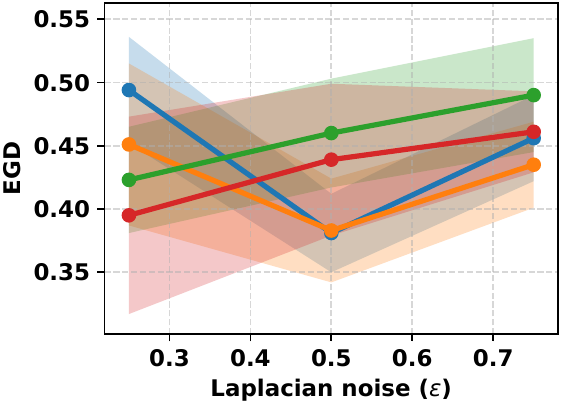}
    
\end{subfigure}
\hfill
\begin{subfigure}{0.23\textwidth}
    \centering
    \includegraphics[width=\linewidth]{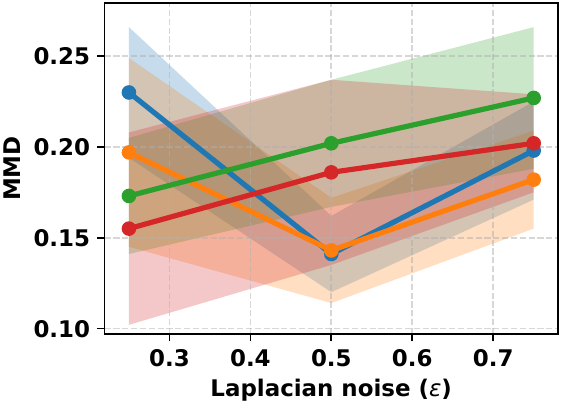}
  
\end{subfigure}
\hfill
\begin{subfigure}{0.23\textwidth}
    \centering
    \includegraphics[width=\linewidth]{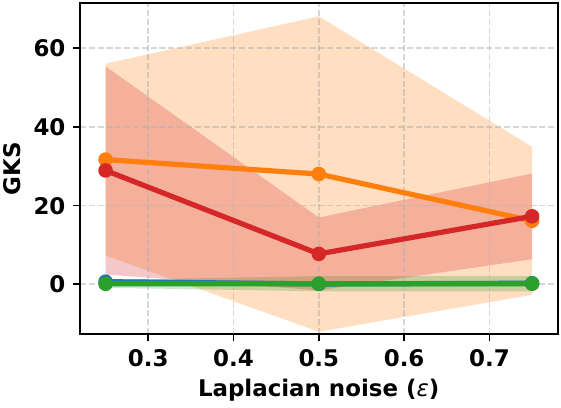}
  
\end{subfigure}

\vspace{2mm}

\textbf{(a) NCI1}

\vspace{3mm}

\begin{subfigure}{0.23\textwidth}
    \centering
    \includegraphics[width=\linewidth]{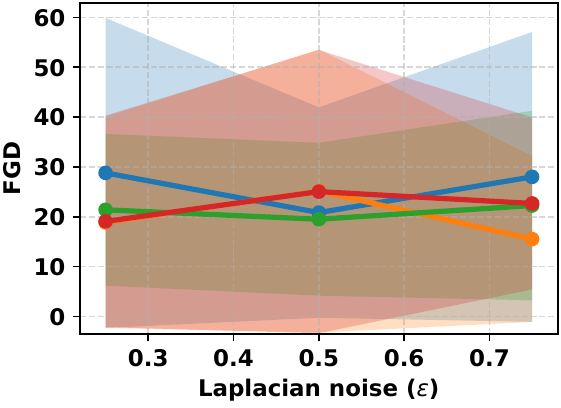}
\end{subfigure}
\hfill
\begin{subfigure}{0.23\textwidth}
    \centering
    \includegraphics[width=\linewidth]{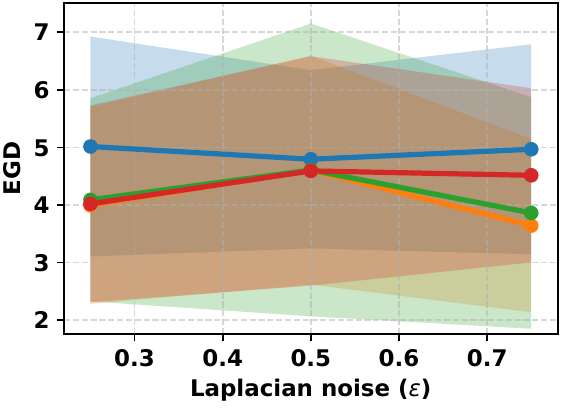}
\end{subfigure}
\hfill
\begin{subfigure}{0.23\textwidth}
    \centering
    \includegraphics[width=\linewidth]{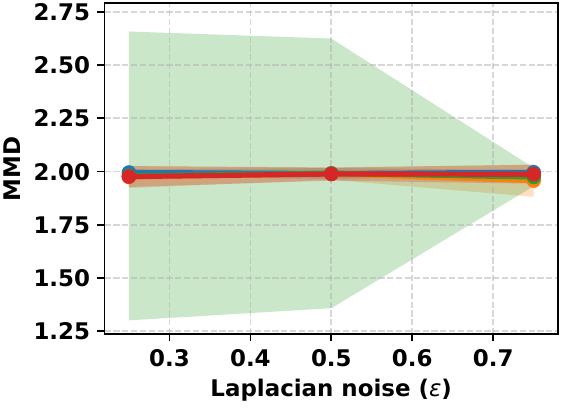}
\end{subfigure}
\hfill
\begin{subfigure}{0.23\textwidth}
    \centering
    \includegraphics[width=\linewidth]{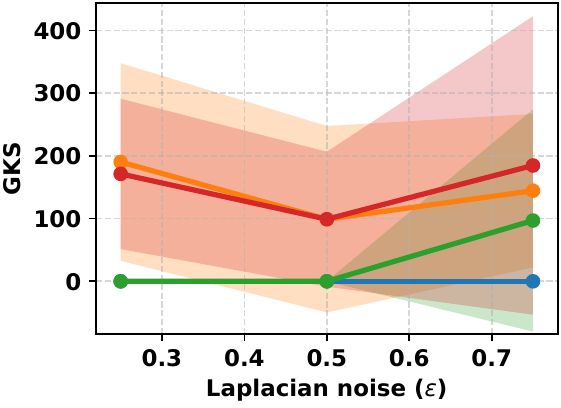}
\end{subfigure}

\vspace{2mm}

\textbf{(b) PROTEINS}

\vspace{3mm}

\begin{subfigure}{0.23\textwidth}
    \centering
    \includegraphics[width=\linewidth]{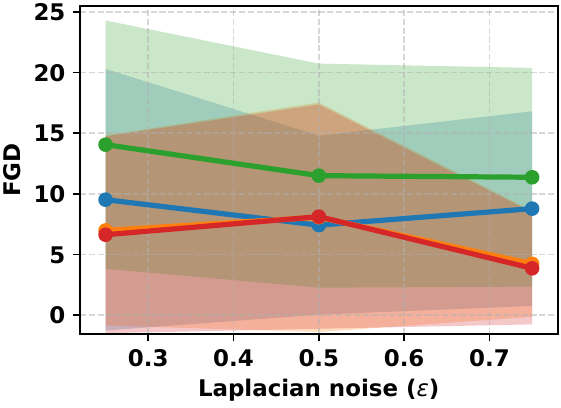}
\end{subfigure}
\hfill
\begin{subfigure}{0.23\textwidth}
    \centering
    \includegraphics[width=\linewidth]{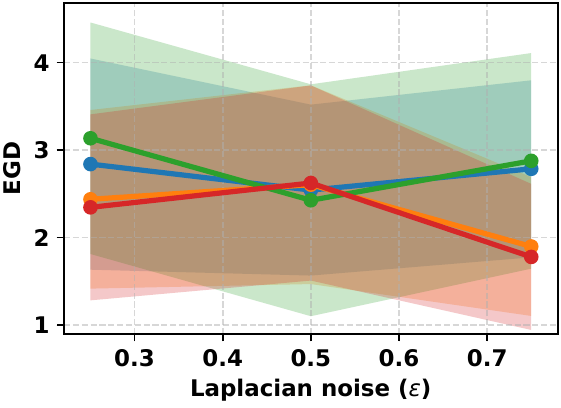}
\end{subfigure}
\hfill
\begin{subfigure}{0.23\textwidth}
    \centering
    \includegraphics[width=\linewidth]{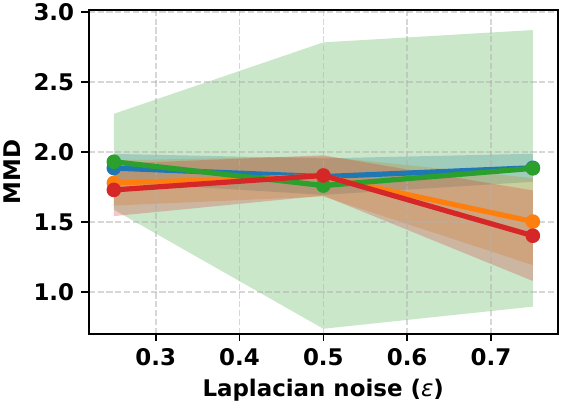}
\end{subfigure}
\hfill
\begin{subfigure}{0.23\textwidth}
    \centering
    \includegraphics[width=\linewidth]{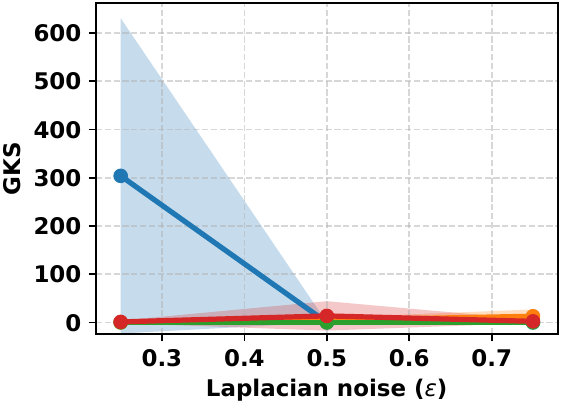}
\end{subfigure}

\textbf{(c) AIDS}
\begin{center}
    \includegraphics[width=0.5\textwidth]{Figures/updated_figure/GLGRA_legend.pdf}
\end{center}
\caption{\textcolor{black}{Comparison of graph reconstruction quality under \textcolor{black}{ELC} attack with varying Laplace scales for three datasets. Columns correspond to FGD, EGD, MMD, and GKS, respectively. Rows correspond to NCI1, PROTEINS, and AIDS datasets.}}
\label{fig:ELC_plot}
\end{figure*}

\textcolor{black}{\subsection{Evaluation Metrics}}
\label{subsec:eval_all}
We applied four metrics to evaluate and compare our model inversion attack against baseline attacks from prior literature.  \\
\textbf{FGD:}
The Fréchet distance in a graph is a similarity measure between paths or walks~\cite{morris2020tudataset}.\\
\textbf{EGD:}
The number of edges on a shortest path from two vertices, u and v, is the graph distance between them~\cite{article}.\\
\textbf{MMD:} This represents a measure of difference between distributions, because of its simplicity, appealing theoretical characteristics, and ease of empirical estimate~\cite{dhanakshirur2024continuous}.\\
\textbf{GKS:}
By computing a symmetric, positive semi-definite function defined on the space of graphs, GKS provides a formal metric for evaluating the similarity between two graphs (like molecular structures)~\cite{schulz2021generalizedweisfeilerlehmangraphkernel}.\\
\section{Experimental Results}
\label{sec:results}
In this section, we use cutting-edge graph reconstruction evaluation criteria to assess our novel attack strategies. We examine our findings and make significant deductions from the results of our experiment.

\subsection{Graph \& Label-Conditioned Graph Reconstruction Attacks }\label{subsec:perform_glgra}
\subsubsection{NCI1}
\textcolor{black}{In both attacks, for Ours method, we train the attack model with 100\% of the dataset, while we train Ours$--$  with 50\% of the dataset, the same applies to the VAE and VAE$--$  during training. The results from our experiments, as shown in Table~\ref{table:merged}, Ours and Ours$--$  attack technique consistently performs better than the attack on the VAE and VAE$--$ model across all metrics, particularly at an $\epsilon$ value of 0.5, where we observe a more stable outcome. With improvements of roughly 48\% in FGD, 31\% in EGD, and 50\% in MMD, our approach specifically produces significant reductions in reconstruction errors, indicating a more accurate recovery of graph distributional properties. The approach retains large increases in capturing graph features, even while performance marginally varies at increasing noise levels ($\epsilon$ = 0.75).}
\subsubsection{PROTEINS}
The result from \textcolor{black}{Table~\ref{table:merged}} illustrates that Ours and Ours$--$ perform better than the VAEs approach. However, it can be deduced from the PROTEINS dataset that the level of noise ($\epsilon$) affects the reconstruction of the graph properties. With an $\epsilon$ value of 0.25, the FGD score of Ours$--$ reduced by 0.3\% when compared to VAE$--$ while Ours has a remarkable reconstruction value of 12\% when compared to the VAE. Ours consistently outperforms VAE for EGD and GKS score except for MMD, with a drop of 0.2\% in reconstruction performance. For $\epsilon$ value of 0.5 and 0.75 Ours and Ours-- outperform their counterpart for all the metrics except EGD, with a drop of 7\% in reconstruction performance. This demonstrates the importance of carefully selecting the appropriate noise value to achieve a very good graph reconstruction outcome.
\subsubsection{AIDS}
The result from empirical analysis of \textcolor{black}{GLC} using the AIDS dataset is shown in Table~\ref{table:merged}. Our and Ours$--$ perform better than the VAEs approach. With an $\epsilon$ value of 0.25, the FGD, EGD, and MMD scores of Ours and Ours$--$ outperform the VAE and VAE$--$. The Ours$--$ performs better than the VAE$--$ with an $\epsilon$ value of 0.5 in two of the metrics, FGD and GKS. While for 0.75 $\epsilon$ value, the Ours and Ours$--$ consistently outperform the VAE and VAE$--$, thus strengthening the importance of $\epsilon$ value selection. It can be observed that the selection of the $\epsilon$ value is also dependent on the nature of the dataset. Due to differences in the sparsity of data, it is mandatory to carefully select an $\epsilon$ value that will enable adequate graph data reconstruction. 
\subsection{Intermediate Rep-Based Graph Reconstruction Attacks }\label{subsec:perform_irgra}
\subsubsection{NCI1}
\textcolor{black}{Table~\ref{table:merged}} shows the result from our empirical analysis of the embedding-based reconstruction attack. In comparison to the VAE-based attack across various $\epsilon$ values, the proposed approach, Ours and Ours$--$, typically offers more efficient graph reconstruction and information extraction. Our method achieves higher distributional similarity at lower noise $\epsilon$ = 0.25, with slightly lower FGD, EGD, and MMD values than the VAEs. It also significantly improves GKS, showing much stronger recovery of graph knowledge. Ours remains competitive with much greater GKS at $\epsilon$ value of 0.50, demonstrating strong attack capacity even under moderate noise. Overall, the experiment shows that the suggested methodology not only reconstructs graph structures more accurately but also pulls significantly more sensitive information from the target model, indicating increased privacy leakage.
\subsubsection{PROTEINS}
The result from Table~\ref{table:merged} illustrates that Ours and Ours$--$ perform better than the VAEs approach. With an $\epsilon$ value of 0.25 and 0.50, there is no visibly wide margib between the performance of each of the techniques across the four metrics. However, with $\epsilon$ value of 0.75, the FGD, EGD, MMD, and GKS score of Ours and Ours$--$ consistently perform better than the VAE and VAE$--$. This illustrate the importance of appropriate noise to dataset selection to achieve a very good graph reconstruction outcome.

\subsubsection{AIDS}
The result from empirical analysis of \textcolor{black}{ELC} using the AIDS dataset is shown in Table~\ref{table:merged}. Ours and Ours$--$ perform better than the VAEs approach. With an $\epsilon$ value of 0.25, the FGD, EGD, MMD, and GKS of Ours outperform the VAE and VAE$--$. While Ours$--$ performs better than the VAE$--$ with $\epsilon$ value of 0.5, thus strengthening the importance of $\epsilon$ value selection. It can be observed that the selection of the $\epsilon$ value is also dependent on the nature of the dataset. Due to differences in the sparsity of data, it is mandatory to carefully select an $\epsilon$ value that will enable adequate graph data reconstruction. Also, the selection of the $\epsilon$ value depends on the dataset sparsity, as this affects the computational and analytical efficiency of the data.

\subsection{Impact of Noise Parameter $\epsilon$}\label{subsec:impact_epsilon}
In our experiment, we applied three (3) variations of the $\epsilon$ value: 0.25, 0.5, and 0.75. It was observed that the attack performance varies with respect to the $\epsilon$ value. With an $\epsilon$ value of 0.25, the performance of the attack degrades, thus affecting the accuracy of the graph reconstruction. The $\epsilon$ value of 0.75 achieves a more stable attack with a high degree of target model training graph reconstruction, while the 0.5 $\epsilon$ value performs slightly better than the result obtained from the $\epsilon$ value of 0.25. In general, $\epsilon$ serves as a noise tuning parameter; excessive or insufficient noise can lower stability, while intermediate values offer the optimal trade-off between stability and reconstruction.

\subsection{Qualitative Comparisons}\label{subsec:qualitative_cmp}
\textcolor{black}{Figures~\ref{fig:GLC_plot} and ~\ref{fig:ELC_plot} show the relationship between our metrics score and Laplacian noise using a line plot. 
Figure~\ref{fig:GLC_plot} shows the result for the \textcolor{black}{GLC} attack method. It can be observed that the NCI1 data consistently display a better score value, as the line graph sags for all the metrics except for the GKS, where higher is better, and the line is clearly visible above other lines. For the PROTEINS and AIDS dataset, the same sagging properties can be observed generally, but not as visibly clear as the NCI1 data. Figure~\ref{fig:ELC_plot} shows the result for the \textcolor{black}{ELC} attack method. The VAE and VAE$--$ line plots are seen to be above all other line plots across almost all the metrics except GKS, where they are consistently seen below Ours and Ours$--$. It can be concluded from the line plots that Ours and Ours$--$  have a better performance when compared to the VAEs. The stability of our approach was also preserved across all four (4) metrics as more of the blue and green shadows are seen overlapping the other shadows. }

\textcolor{black}{\section{Discussion and Future Work}
\label{sec:Discuss}
\subsection{Discussion}
This study proposes two attack approaches, \textcolor{black}{GLC} and \textcolor{black}{ELC}, for training graph data reconstruction of GNN models with two attack variants, Ours and Ours$--$. Ours involves training the attack model with 100\% of the dataset, Ours$--$ with 50\% of the dataset. From the empirical analysis carried out using different datasets for model inversion on the GNN model, we deduced that the underlying relationship between model output and its training data makes GNNs highly susceptible to privacy attacks. This vulnerability is not enough for the adversary to successfully reconstruct the GNN model training data. However, the attack model architecture and the knowledge of the shadow dataset ($\epsilon$ value selection) can affect the computational and analytical efficiency to successfully reconstruct GNN training data. By characterizing an inversion attack as a conditional graph generation problem utilizing a GAN-based model, leaked embeddings, and prediction from the victim model, we quantify the leaking of GNN models.
\subsection{Future Work}
The graph data reconstruction opens up privacy threat analysis in IoT and network-connected domains, including biometric data leakage~\cite{vhaduri2022predicting} and intrusion detection~\cite{mukherjee1994network} with graph-based data representations. Future study in MI attack is to develop a suitable defense  technique that would negatively impact the high correlation between model prediction and model training data without compromising the utility of the model. To improve MI attack, future studies will consider how to invert GNN models with low correlation between model prediction and the data the model is trained on, a study area considered to be known as a query-free attack. The research should also focus on expanding real-life applications, e.g., predicting sensitive user patterns or lifestyles, based on connectivity and privacy-sensitive network information like phone call patterns~\cite{vhaduri2021deriving}.}
\section{Conclusion}
\label{sec:conclusion}
\textcolor{black}{We present two attacks, \textcolor{black}{GLC} and \textcolor{black}{ELC}, for GNNs training data reconstruction using a GAN model. We vary $\epsilon$ values to compare the reconstruction performance of our GAN-based methods with baselines across our two proposed attacks. Ours and Ours$--$ at $\epsilon$ of 0.75 demonstrate a much higher attack stability and reconstruction ability across the four (4) metrics. The observed results show that the proposed attack approach using a GAN model is significantly more effective than VAE methods, demonstrating a higher degree of privacy leakage in the GNN model. This research sheds light on future research for stronger defense design on graph reconstruction attacks, benefiting trustworthy AI research.}

\section*{Acknowledgment}

We are grateful for the generous support from Google for the research. We want to thank the anonymous reviewers for their helpful feedback.

\bibliographystyle{ieeetr}
\bibliography{ref}


\end{document}